\documentclass[11pt]{article}
\usepackage{acl}
\usepackage{times}

\usepackage[utf8]{inputenc}
\usepackage[T1]{fontenc}
\usepackage{latexsym}
\usepackage{url}
\usepackage{booktabs}
\usepackage{multirow}
\usepackage{amsfonts}
\usepackage{amsmath}
\usepackage{nicefrac}
\usepackage{microtype}
\usepackage{inconsolata}
\usepackage{xcolor}
\usepackage{colortbl}
\usepackage{graphicx}
\usepackage{enumitem}
\usepackage{pifont}
\usepackage{cleveref}

\crefname{figure}{Figure}{Figures}
\crefname{table}{Table}{Tables}
\crefname{section}{Section}{Sections}
\crefname{appendix}{Appendix}{Appendices}

\newcommand{\cmark}{\ding{51}}
\newcommand{\xmark}{\ding{55}}

\title{What Happens Inside Agent Memory?\\Circuit Analysis from Emergence to Diagnosis}

\author{Xutao Mao \\
City University of Hong Kong \\
\And
Jinman Zhao \\
University of Toronto \\
\AND
Gerald Penn \\
University of Toronto \\
\And
Cong Wang \\
City University of Hong Kong \\
}

\begin{document}

\maketitle

\begin{abstract}
Agent memory failures are silent: an LLM-based agent can produce a fluent response even when it fails to extract, retain, or retrieve the information needed across sessions. The \emph{write-manage-read} loop describes the external pipeline of these systems but leaves open which internal computations implement each stage. We trace feature circuits across the Qwen-3 family (0.6B--14B), two memory frameworks (mem0, A-MEM), and a matched Gemma-3 replication, reporting two mechanistic findings plus one deliverable. First, under mem0, control is detectable before content. Routing circuitry is causally active at 0.6B, while content circuitry produces no detectable signal until 4B. Gemma-3 reproduces the same order at 1B and 4B. Small models therefore route memory decisions before they can reliably extract or ground the underlying facts. Second, the shared hub is recruited, not created. Write and Read converge on a late-layer hub that already exists in the base model as a context-grounding substrate, on which memory framing recruits a memory-specific functional direction. The hub and the early routing circuit carry over to A-MEM, indicating that these computations belong to the base model rather than to a particular interface. Building on this circuit structure, we develop an unsupervised stage-level diagnostic that localizes silent failures to the responsible operation at 76.2\% accuracy. It outperforms a strong LLM trace judge by 3.1 points as well as a supervised hidden-state probe by 8.8 points. Together, these results establish circuit-level signatures as a practical handle for monitoring agent memory and guiding its structural design. 

\end{abstract}

\vspace{-2mm}
\section{Introduction}
\label{sec:intro}

\begin{figure}[t]
    \centering
    \includegraphics[width=0.64\columnwidth]{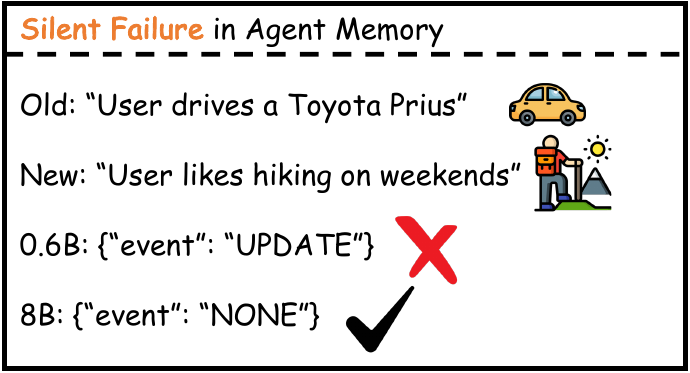}
    \caption{Silent Manage failure in Qwen-3: 0.6B overwrites a car fact with an unrelated hiking preference (UPDATE), whereas 8B preserves it (NONE).}
    \label{fig:silent-failure}
\vspace{-4mm}
\end{figure}

Agent memory systems enable LLM-based agents to persist, organize, and recall information across sessions. They have become a core component of production agent architectures~\citep{chase2022langchain,chadha2025mem0}. \citet{zhang2025survey,du2026memory} formalize the shared structure underlying these systems as a write-manage-read loop. The agent extracts facts from conversations (Write), decides whether to add, update, delete, or keep stored memories (Manage), then grounds its answers in retrieved content (Read). This loop unifies mem0~\citep{chadha2025mem0}, A-MEM~\citep{amem2025}, and RL-trained systems~\citep{yan2026memoryr1,huo2026atommem}.

The decomposition suggests a natural diagnostic question: when an agent gives a wrong answer, which stage failed? External behavior cannot answer it. Extracted facts look well-formed, routing decisions are legal tokens, and final answers are fluent, so failures are silent by construction (\cref{fig:silent-failure}). Evaluation collapses extraction, routing, and grounding errors into one end-to-end accuracy~\citep{cemri2025why,hu2026memoryagentbench}. As these systems take on longer horizons, undiagnosed failures compound~\citep{zhao2026ama}.

Mechanistic interpretability has addressed related problems in single-turn retrieval. Within a single forward pass, ReDeEP dissects RAG hallucination~\citep{sun2025redeep}, Retrieval Heads locate context retrieval~\citep{wu2025retrievalhead}, and ROME/MEMIT localize factual associations in MLPs~\citep{meng2022locating,meng2023massediting}. But agent memory is not a single call. A Write failure corrupts what enters the store, while a wrong Manage decision persists in it, yet both surface only at Read time. Each stage is a separate forward pass, but no existing work traces circuits through this multi-call pipeline~\citep{hu2025memory}. Scale is also unaddressed. Agent memory systems are deployed across a wide range of model sizes, where the same framework can behave differently depending on the backbone~\citep{bandel2026general,zhang2026actmem}. Only cross-scale internal analysis can tell a capacity limit from a structural bottleneck~\citep{nanda2023progress,bahri2024explaining}.

This paper investigates whether the write--manage--read stages of agent memory leave separable internal signatures that are stable enough to localize silent failures to specific pipeline stages. To this end, we trace feature circuits~\citep{ameisen2025circuittracer} through each memory operation across the Qwen-3 family (0.6B, 4B, 8B, 14B), two memory frameworks (mem0, A-MEM), and a matched replication on Gemma-3 (1B, 4B). To our knowledge, this is the first circuit-level analysis of a multi-call agent memory pipeline. Our contributions are:
\begin{figure*}[t]
    \centering
    \includegraphics[width=0.56\textwidth]{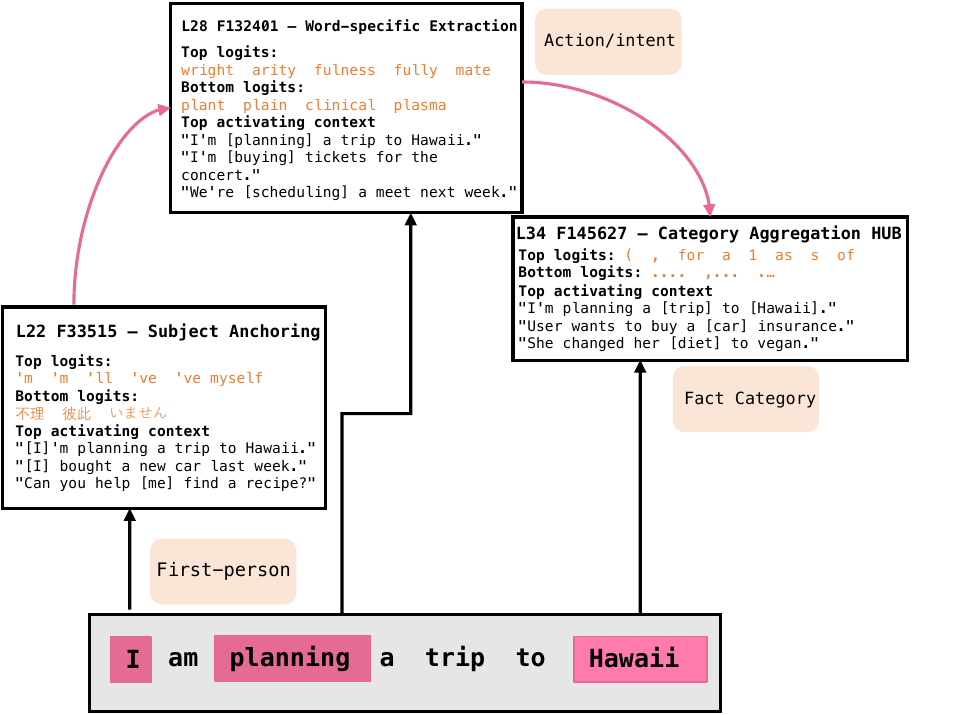}
    \caption{Qwen-3 8B Write circuit for ``I'm planning a trip to Hawaii.'' Arrows trace information flow from input tokens through subject anchoring (L22), word-specific extraction (L28), and category aggregation (L34). Boxes show feature IDs, top/bottom logits, and activating contexts that support the feature interpretations.}
    \label{fig:hero}
\vspace{-4mm}
\end{figure*}

\begin{enumerate}[itemsep=1pt,topsep=3pt]
    \item \emph{Control-before-content asymmetry.} Under mem0, routing circuitry (Manage) is causal at 0.6B, while content circuitry (Write, Read) produces no detectable signal until 4B. Small agents thus route memories before they can extract or ground them. The order replicates on Gemma-3 (1B, 4B) under the same prompts. Under A-MEM, Manage is again causal at 0.6B, though its structured Write is too.

    \item \emph{A hub recruited, not created.} Write and Read converge on a late-layer hub that already exists in the base model as a generic context-grounding substrate. Memory framing does not build new machinery but recruits a memory-specific functional direction on it. The hub also carries over to A-MEM. Reliability is thus bounded by what the base model already supports.

    \item \emph{Unsupervised stage-level diagnostic.} We turn the feature-space separation between routing and content circuits into a zero-shot diagnostic that localizes silent failures to the responsible operation at 76.2\% accuracy. It outperforms a zero-shot LLM trace judge (73.1\%) as well as a supervised hidden-state probe (67.4\%), and it transfers across benchmarks without retraining. Because amplifying content circuits proves fragile in the pipeline, the diagnostic reads circuits out rather than steering them.
\end{enumerate}

\vspace{-2mm}
\section{Background and Related Work}
\label{sec:background}

\paragraph{Agent Memory Systems and Where They Fail.}

\citet{zhang2025survey,du2026memory} unify the design space of agent memory under a write-manage-read loop. It spans explicit LLM calls~\citep{chadha2025mem0,amem2025}, hierarchical paging~\citep{packer2024memgpt}, observation-reflection-retrieval~\citep{park2023generative}, RL-trained policies~\citep{yan2026memoryr1,huo2026atommem,yu2026agemem}, and graph-based storage~\citep{magma2026}. Memory operations shape long-term agent behavior~\citep{xiong2025howmemory}. Benchmarks expose persistent weaknesses such as selective forgetting~\citep{hu2026memoryagentbench} or a gap between passive recall and active use~\citep{he2026memoryarena}. When such systems fail, Who\&When~\citep{zhang2025which} and AgenTracer~\citep{zhang2026agentracer} attribute each failure to an agent or step from external behavior. These methods observe only what the pipeline emits. Circuit signatures instead read the internal computation directly, which matters precisely when the emitted behavior is fluent but wrong.

\paragraph{Circuits, Retrieval, and Error Detection.}

Circuit tracing with transcoders~\citep{dunefsky2024transcoders,ameisen2025circuittracer} produces feature-level attribution graphs. Sparse feature circuits~\citep{marks2025sparse,kharlapenko2025scaling} extend it to practical models, while cross-task reuse studies~\citep{merullo2024reuse,mondorf2025compositions,he2025modcirc} examine how circuits are shared. Most relevant, \citet{hanna2026planning} trace planning circuits across Qwen-3, finding scale-dependent emergence. Each edge is an exact direct effect under a fixed linearization, so feature interventions can test a traced graph, which also makes graphs comparable across stages and scales. Within single calls, ReDeEP~\citep{sun2025redeep} dissects RAG hallucination, Retrieval Heads~\citep{wu2025retrievalhead} locate context retrieval, ContextFocus~\citep{anand2026contextfocus} steers faithfulness, RAGLens~\citep{xiong2025raglens} detects unfaithful outputs, and ROME/MEMIT localize factual associations in MLPs~\citep{meng2022locating,meng2023massediting,geva2023dissecting}. Hidden states further reveal whether an output is wrong, via truthfulness probes~\citep{azaria-mitchell-2023-internal}, semantic entropy~\citep{DBLP:journals/nature/FarquharKKG24}, and single-pass approximations thereof~\citep{han2024semantic}. We extend both lines from the single call to the multi-call write-manage-read loop, tracing how memory circuits develop with scale while attributing a silent failure to the operation that caused it.

\vspace{-2mm}
\section{Method}
\label{sec:method}

\vspace{-2mm}
\subsection{Agent Memory Pipeline}
The pipeline comprises three LLM operations: Write (extract facts from conversation), Manage (decide add/update/delete/none), and Read (answer from retrieved memory). Each is an independent forward pass with a stage-specific prompt. The pipeline processes sessions sequentially. Each session triggers Write, which adds facts to the store, followed by Manage, which compares each new fact against its top-5 retrieved neighbors to decide add/update/delete/none per pair. At question time, the top-5 facts are retrieved for Read. In both mem0 and A-MEM, retrieval uses embedding-based search without an LLM call. Top-5 matches the evaluated mem0 configuration, and \cref{app:retrieval-sensitivity} re-runs the full analysis at retrieval depths $1$--$10$, finding that the stage signatures persist.

Our primary system is mem0~\citep{chadha2025mem0},\footnote{Our analysis targets the pre-v2.0 mem0 architecture.} a widely adopted open-source agent memory framework that stores extracted facts as flat key-value entries and retrieves them via semantic search. We additionally analyze A-MEM~\citep{amem2025}, which organizes memories as a self-linking Zettelkasten network. Following the released implementation, A-MEM issues two LLM calls per memory event, namely Note Construction (Write) plus a single Evolution call that jointly handles link generation and neighbor-context updates (Manage). We also construct a single-operation Read prompt that consumes A-MEM's memory triples, then trace circuits through each stage (\cref{app:amem-prompts}). Comparing circuits across the two systems tests whether memory circuits persist or are interface-specific. Both systems are evaluated on LongMemEval~\citep{wu2025longmemeval}, a multi-session chat-assistant memory benchmark. Each of its 500 questions embeds 1--4 answer-bearing sessions in a long history of filler sessions, covering single-session recall, cross-session recall, temporal reasoning, knowledge updates, and multi-hop questions (\cref{app:eval-dataset}).

\vspace{-2mm}
\subsection{Feature Circuits with Pre-Trained Transcoders}

We trace feature circuits~\citep{ameisen2025circuittracer} through each memory operation. Per-layer transcoders (PLTs)~\citep{dunefsky2024transcoders} replace each MLP layer with a sparse encoder-decoder that produces monosemantic feature activations. The resulting feature circuit connects input token embeddings, transcoder features (identified by layer, position, feature index), and output logit nodes through weighted causal edges. These edges are computed exactly under a locally linear replacement model in which attention patterns and normalization terms are frozen. Attention mixes information across token positions, while MLP blocks transform the residual content at each position~\citep{wang2023ioi,olsson2022induction}. Tracing the MLP pathway therefore captures how routed memory information is transformed, with attention held fixed as a controlled estimand. The \texttt{mwhanna} PLT collection supports this design by providing matched feature interventions at every MLP layer of all four Qwen-3 scales. Throughout the paper, ``no detectable signal'' means no detectable causal contribution within this traced MLP pathway. All circuits are traced on correct instances, filtered by per-stage correctness flags assigned by a Qwen-3 32B judge (\cref{app:eval-dataset}), so the discovered circuits reflect successful computation rather than error pathways.

For each operation at each scale, we apply a four-step pipeline (hyperparameters in \cref{app:hyperparams}):
Step~1 (Attribution). Per-sample feature circuits via backward pass through the replacement model, retaining the top 4,096 feature nodes and pruning to 80\% node influence / 95\% edge influence.
Step~2 (Aggregation). Cross-sample recurrence counting, reducing each circuit to 50--200 recurrent features.
Step~3 (Path tracing). Canonical paths extracted by greedily following highest-weighted edges from input to output.
Step~4 (Causal verification). On a held-out 30\% split~\citep{hanna2026planning}, we compare zero ablation as well as $5\times$ amplification of top-10 circuit features against matched random baselines.\footnote{Random baseline features are drawn uniformly from active features at the same layers and token position outside the top-$k$ circuit. Each of 5 draws uses a distinct set, with results averaged.} The target $y^*$ is the model's own top prediction at the JSON prefix position (up to 5 logits covering $\geq$80\% cumulative probability, \cref{app:method}). The primary metric is $\overline{\Delta p}$, the mean change in probability of $y^*$ under intervention. The causal gap, $|\overline{\Delta p}_{\text{circuit}} - \overline{\Delta p}_{\text{random}}|$, isolates discovered features from arbitrary same-layer features. We report causal gaps with bootstrap 95\% CIs (\cref{app:bootstrap}). A near-zero gap means no contribution is detected. We also report Jaccard similarity $J(A,B) = |A \cap B| / |A \cup B|$ over top-30 features to quantify circuit sharing.

\vspace{-2mm}
\subsection{Experimental Setup}
\label{sec:models}
Each stage uses a single-operation prompt ending with a JSON prefix that positions the model to emit the stage-specific content token next: an extracted fact after \texttt{\{\textquotedbl facts\textquotedbl:\textquotedbl User} (Write), one of ADD/UPDATE/DELETE/NONE after \texttt{\{\textquotedbl event\textquotedbl:\textquotedbl} (Manage), or an answer after \texttt{\{\textquotedbl answer\textquotedbl:\textquotedbl} (Read). Full templates are in \cref{tab:mem0-prompts,tab:amem-prompts}. Single-operation prompts hold operation semantics fixed along with history, making circuits comparable across stages and scales. We study four models from the Qwen-3 family (0.6B, 4B, 8B, 14B)~\citep{qwen2025}, each with corresponding PLTs from the \texttt{mwhanna} transcoder collection ($d_{\text{feature}} = 163{,}840$, ReLU, full-layer coverage). For cross-family replication (\cref{sec:crossfamily}), we run the matched protocol on Gemma-3 IT (1B, 4B)~\citep{team2025gemma} with gemma-scope-2 transcoders~\citep{gemmascope2}. For cross-system analysis (\cref{sec:system}), we replace mem0's prompts with A-MEM's equivalents. All experiments run on H200 GPUs (\cref{app:hyperparams}).

\vspace{-2mm}
\section{Control Before Content: Which Stage Circuits Emerge First?}
\label{sec:results}

\begin{figure*}[t]
    \centering
    \includegraphics[width=0.62\textwidth,trim={0 0 0 30pt},clip]{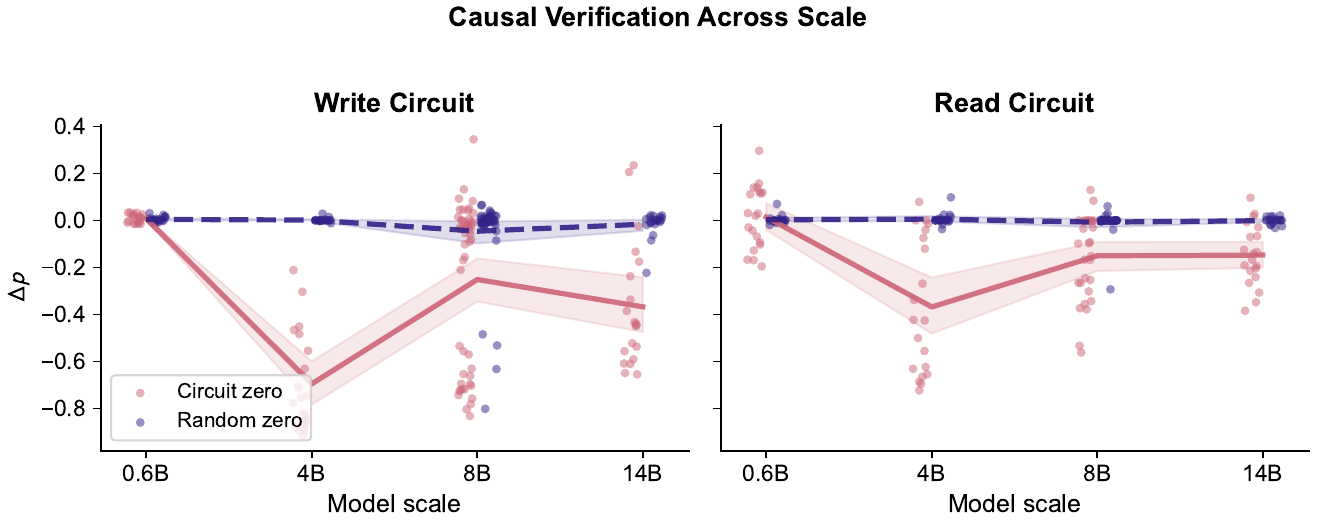}\hspace{0.02\textwidth}
    \includegraphics[width=0.29\textwidth]{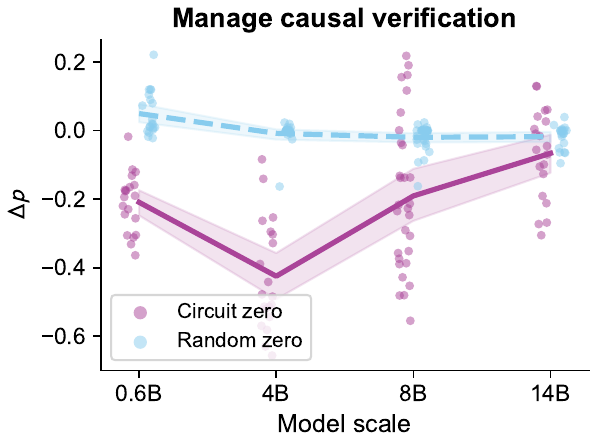}
    \caption{Causal verification across Qwen-3 scales. Zero ablation of the top-10 circuit features (solid) is compared with layer-matched random features (dashed) by the mean change in target-token probability. Points are held-out samples, with bootstrap 95\% CIs as bands. Manage (right) separates from its random baseline at 0.6B, whereas Write and Read (left, center) do so only from 4B. All three gaps peak at 4B.}
    \label{fig:causal-verification}
\vspace{-4mm}
\end{figure*}

We apply the four-step analysis pipeline (\cref{sec:method}) with mem0 prompts to each memory operation at four Qwen-3 scales and two Gemma-3 scales, producing 200 feature circuits per operation per model. \cref{tab:summary-main} summarizes all models, while \cref{fig:causal-verification} shows the per-sample interventions behind the Qwen-3 gaps. The organizing finding is a control-before-content asymmetry. Routing circuitry (Manage) produces a detectable causal gap at the smallest scale of both families, whereas content circuitry (Write, Read) produces none until 4B.

\begin{table*}[t]
\centering
\caption{Stage circuits across scales and model families. Gaps are causal gaps $|\overline{\Delta p}_{\text{circuit}}-\overline{\Delta p}_{\text{random}}|$ for top-10 zero ablation on the held-out 30\% split, with bootstrap 95\% CIs for Qwen-3 (\cref{app:bootstrap}). \colorbox{gray!20}{Shaded} cells have CIs that include zero. $J$ is top-30 feature Jaccard overlap between stages. Hub is the late-layer cluster shared by Write and Read. Routing is detectable at the smallest scale of both families, whereas content circuits appear together with the shared hub at 4B.}
\label{tab:summary-main}
\footnotesize
\setlength{\tabcolsep}{4pt}
\begin{tabular*}{\textwidth}{@{\extracolsep{\fill}}lccccccc@{}}
\toprule
& & \multicolumn{3}{c}{\textbf{Causal gap [95\% CI]}} & \multicolumn{2}{c}{\textbf{Overlap}} & \\
\cmidrule(lr){3-5} \cmidrule(lr){6-7}
\textbf{Model} & \textbf{Layers} & \textbf{Write} & \textbf{Manage} & \textbf{Read} & $J$(W,R) & $J$(W,M) & \textbf{Hub} \\
\midrule
Qwen-3 0.6B & 28 & \cellcolor{gray!20}.004 [$-$.003, .011] & .259 [.198, .321] & \cellcolor{gray!20}.012 [$-$.005, .029] & .132 & .071 & none \\
Qwen-3 4B   & 36 & .697 [.601, .789] & .417 [.341, .492] & .374 [.264, .484] & .154 & .071 & L35 \\
Qwen-3 8B   & 36 & .224 [.131, .318] & .172 [.108, .237] & .142 [.085, .200] & .111 & .053 & L34 \\
Qwen-3 14B  & 40 & .351 [.271, .432] & .048 [.021, .075] & .146 [.094, .198] & .091 & .000 & L38--39 \\
\midrule
Gemma-3 1B  & 26 & .06 & .27 & .05 & .11 & -- & none \\
Gemma-3 4B  & 34 & .29 & .41 & .33 & .277 & -- & L31 \\
\bottomrule
\end{tabular*}
\vspace{-4mm}
\end{table*}

\vspace{-2mm}
\subsection{Routing Circuits Are Detectable Early}
\label{sec:manage}

The Manage circuit produces a significant causal gap at 0.6B ($.259$, CI $[.198,.321]$), the only operation with detectable signal at this scale (\cref{tab:summary-main}). Conflict detection lives in shallow layers available even at the smallest model: decision-type features appear as early as L16 at 0.6B.

The Manage circuit exhibits a trunk-plus-routing architecture at all scales where it is causal. At 8B, its canonical path is L14$\to$L18$\to$L35. The trunk (L13--18) processes the semantic content of the memory conflict, with features that fire on user-content tokens regardless of the eventual decision. The routing stage (L19--28) then selects among the four output actions, with each action recruiting a distinct feature set (update--delete Jaccard $.558$, \cref{app:manage}). Its action-associated features are confirmed on held-out graphs at F1 $.75$--$.82$ (\cref{app:feature-validation}). The same two-stage organization appears at every scale, with both stages migrating to deeper layers as models grow. Early maturity has a cost. At small scales, routing circuitry is active before the model holds a robust representation of the content it routes, so memory management proceeds on unreliable decisions (\cref{fig:silent-failure}).

\vspace{-2mm}
\subsection{Content Circuits Emerge Late and Distribute with Scale}
\label{sec:write}

Write and Read show no detectable causal signal at 0.6B (CIs include zero), despite $.340$ intact QA accuracy (\cref{tab:sweep}) as well as $98.7\%$ top-1 fidelity of the replacement model (\cref{app:fidelity}). Features at this scale spread across ten layers with no concentration. Read features fire only on format tokens rather than memory positions (\cref{app:write,app:read}). Both become causal at 4B, where a late-layer hub appears (\cref{tab:summary-main}).\footnote{The hub is a cluster of co-recurring features in the deepest active layers. At 8B, $\sim$10 features near L34 form this cluster, with F145627 as the most recurrent. It shifts from L32--35 at 4B to L37--39 at 14B.}

We illustrate content circuits with Write at 8B (\cref{fig:hero}). Its 200 attribution graphs follow three stages along the canonical path L22$\to$L28$\to$L34$\to$L35: subject anchoring at L22, word-specific extraction at L28, then category aggregation at the L34 hub. The hub appears in 45 of 60 held-out graphs, connecting 96 distinct content-token types. Read follows the same late-layer path. Memory-position features account for 62\% of its attribution, while ablating the L34 hub reduces the answer probability $\sim$47$\times$ more than a random L34 feature (\cref{app:read}).

The Write gap peaks at 4B ($.697$) and declines at 8B--14B. This could reflect weaker circuitry or computation spread beyond the fixed top-10 bank. An ablation-budget sweep supports distribution: from $k{=}10$ to $k{=}100$, the 4B gap stays within $.697$--$.729$, whereas the 8B gap rises from $.224$ to $.358$ and the 14B gap reaches $.426$ at $k{=}50$ (\cref{tab:budget-sweep}). Larger models therefore spread content computation across more features rather than lose it.

\begin{table}[t]
\centering
\caption{Write causal gap as the ablation budget grows from the top-5 to the top-100 recurrent features. The 4B gap saturates by $k{=}20$, while the 8B and 14B gaps keep growing, so larger models spread Write computation over more features.}
\label{tab:budget-sweep}
\footnotesize
\setlength{\tabcolsep}{4pt}
\begin{tabular*}{\columnwidth}{@{\extracolsep{\fill}}lccccc@{}}
\toprule
\textbf{Scale} & $k{=}5$ & $10$ & $20$ & $50$ & $100$ \\
\midrule
4B  & .544 & .697 & \textbf{.729} & .702 & .721 \\
8B  & .131 & .224 & .260 & .297 & \textbf{.358} \\
14B & .287 & .351 & .383 & \textbf{.426} & .410 \\
\bottomrule
\end{tabular*}
\vspace{-5.5mm}
\end{table}

\vspace{-3mm}
\subsection{Two Circuit Groups Diverge with Scale}
\label{sec:cross}

The two groups also separate in feature space. Write--Read overlap exceeds Write--Manage at every scale, with the gap between them widening (\cref{tab:summary-main}, heatmaps in \cref{fig:jaccard-heatmaps-scale}). It grows from $.132$ vs.\ $.071$ at 0.6B to $.091$ vs.\ $.000$ at 14B, where Write shares no top-30 feature with Manage but still shares the hub with Read (L38\,F7190, L37\,F103617). At 8B, 27 of Write's top-30 features lie in L31--35, as do 26 of Read's but only 17 of Manage's. The hub feature L34\,F145627 never enters Manage (\cref{app:cross}). Routing thus occupies a computational space detectable at 0.6B, separate from the content space that forms at 4B.

\vspace{-2mm}
\subsection{The Detection Order Replicates Across Model Families}
\label{sec:crossfamily}

The results above come from one model family, so the detection order could in principle be a property of Qwen-3's pre-training. We rerun the full pipeline on Gemma-3 IT (1B, 4B) with gemma-scope-2 transcoders under the matched protocol, keeping the operation prompts, 70/30 split, correct-only tracing rule, top-10 bank budget, target-token definition, and layer-matched random ablations fixed. Feature banks are rebuilt on each model's own discovery split (\cref{app:gemma}). Gemma-3 reproduces the order (\cref{tab:summary-main}). At 1B, Manage yields a $.27$ gap against $.06$/$.05$ for Write/Read, with activity diffuse across L16--25. At 4B, Write/Read rise to $.29$/$.33$, their overlap grows from $.11$ to $.277$, and they share an L31 hub. Under mem0 prompts, control before content is therefore a cross-family phenomenon. In both families, the hub emerges at the scale where content circuits become detectable.

\vspace{-2mm}
\section{The Shared Hub Is Recruited, Not Created}
\label{sec:hub}

The content circuits of \cref{sec:results} converge on a shared late-layer hub. We hypothesize that this cluster is a generic context-grounding substrate the base model already uses to ground decoding in supplied context. Memory framing, in turn, would recruit a functionally distinct direction on it. Single-pass context-faithfulness work supports the existence of such a substrate. ReDeEP attributes RAG hallucination to context-copying heads as well as parametric-knowledge FFNs~\citep{sun2025redeep}, while RAGLens detects unfaithful RAG outputs with sparse features~\citep{xiong2025raglens}.

\begin{table}[t]
\centering
\caption{Hub tests. (a) Donor-to-recipient transplant of top-10 hub activations: changed top-1 predictions (of 200) with the drop in the recipient's own-answer probability (pp). Layer-matched controls change nothing. (b) Steering conflict prompts by $\beta$ along the memory-grounding direction, where $\Delta M$ is the shift in margin toward the memory answer. (c) 8B steering ($\beta{=}+1$) with directions from memory vs.\ direct-context framing.}
\label{tab:hub-evidence}
\footnotesize
\setlength{\tabcolsep}{3pt}
\begin{tabular*}{\columnwidth}{@{\extracolsep{\fill}}lccccc@{}}
\multicolumn{6}{@{}l}{\textbf{(a) Transport}} \\
\toprule
& \multicolumn{3}{c}{Read$\to$Read} & \multicolumn{2}{c}{8B cross-stage} \\
\cmidrule(lr){2-4} \cmidrule(lr){5-6}
& 4B & 8B & 14B & W$\to$R & R$\to$W \\
\midrule
Changed & 0 & \textbf{110} & 10 & 20 & \textbf{185} \\
Prob.\ drop & $-1$ & $\mathbf{-47}$ & $-1$ & $-18$ & $\mathbf{-72}$ \\
\bottomrule
\end{tabular*}

\vspace{0.5ex}
\begin{tabular*}{\columnwidth}{@{\extracolsep{\fill}}lcccccc@{}}
\multicolumn{7}{@{}l}{\textbf{(b) Engagement $\Delta M$ vs.\ $\beta$}} \\
\toprule
& $-2$ & $-1$ & $-.5$ & $+.5$ & $+1$ & $+2$ \\
\midrule
4B  & $-1.53$ & $-0.41$ & $-0.08$ & $-0.11$ & $-0.42$ & $-1.39$ \\
8B  & $-0.85$ & $-0.53$ & $-0.28$ & $+0.13$ & $+0.10$ & $-0.09$ \\
14B & $-0.84$ & $-0.43$ & $-0.20$ & $+0.16$ & $+0.30$ & $+0.40$ \\
\bottomrule
\end{tabular*}

\vspace{0.5ex}
\begin{tabular*}{\columnwidth}{@{\extracolsep{\fill}}lcc@{}}
\multicolumn{3}{@{}l}{\textbf{(c) Specificity $\Delta M$ (direction $\to$ prompts)}} \\
\toprule
& Memory prompts & Direct-context prompts \\
\midrule
$d_{\text{mem}}$ & $\mathbf{-2.30}$ & $+1.02$ (wrong sign) \\
$d_{\text{dir}}$ & $-0.02$ & $+0.32$ \\
\bottomrule
\end{tabular*}
\vspace{-4mm}
\end{table}

\vspace{-2mm}
\subsection{The Hub Is a Grounding Substrate with a Memory-Specific Direction}
\label{sec:read}

Write and Read converge on the 8B hub despite disjoint output vocabularies (verbs vs.\ nouns, zero token overlap), indicating shared computation rather than output-level similarity. Three tests characterize what the hub carries (\cref{tab:hub-evidence}).

\vspace{-1mm}
\paragraph{It carries sample-specific state, not answers.} Transplanting hub activations between unrelated 8B samples changes 110/200 within-stage predictions, rising to 185/200 for Read$\to$Write, while layer-matched controls change none (\cref{tab:hub-evidence}a). The recipient never switches to the donor's answer, so the hub encodes a context-dependent grounding state rather than a literal value. The effect is scale-specific. At 4B the hub is structurally important but not yet content-sensitive, whereas at 14B the computation is too distributed for ten features to cover (\cref{app:transport}).

\vspace{-1mm}
\paragraph{It controls reliance on memory.} We define a memory-grounding direction as the hub activation difference between memory-present versus memory-absent prompts. We then steer conflict prompts, where memory contradicts parametric knowledge, along this direction. Dampening moves 8B predictions away from the memory answer. The effect becomes cleanly bidirectional at 14B, whereas at 4B any perturbation is destructive (\cref{tab:hub-evidence}b, \cref{app:engagement}).

\vspace{-1mm}
\paragraph{The functional direction is specific to memory.} A direct-context control supplies the same questions with the same facts as a short context block without memory framing (\cref{app:hub-specificity}). The two extracted directions share substantial subspace (cosine $.736$ vs.\ a $.942$ split-half bound), yet only the memory direction is a functional axis. Self-steering yields $\Delta M=-2.30$ under memory framing but only $+0.32$ under direct context. Applied to direct-context prompts, the memory direction even has the wrong sign (\cref{tab:hub-evidence}c). The hub is thus a common substrate on which memory framing recruits its own grounding axis. Memory frameworks do not build grounding machinery. They borrow it, so their reliability is bounded by how well a framing aligns with what the base model already supports.

\vspace{-2mm}
\subsection{Transfer Across Memory Frameworks}
\label{sec:system}

\begin{figure}[t]
    \centering
    \includegraphics[width=0.92\linewidth]{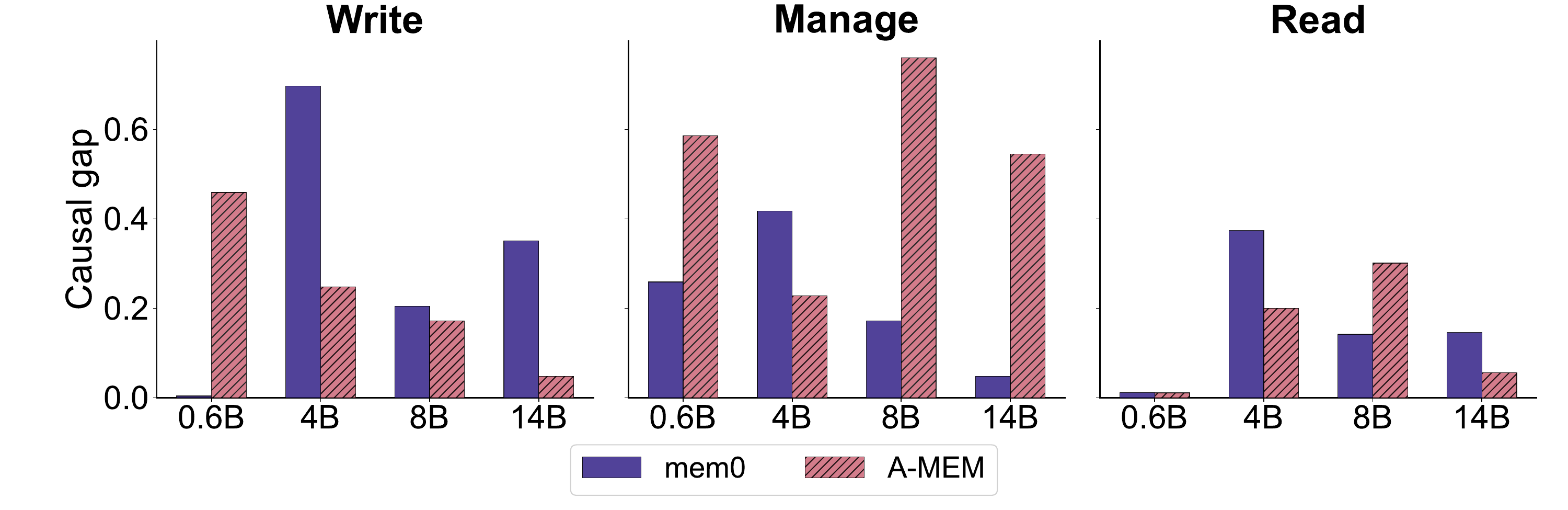}
    \caption{Causal gaps across Qwen-3 scales under mem0 (blue) and A-MEM (red). Manage is causal at 0.6B under both frameworks, as is A-MEM's structured Write.}
    \label{fig:cross-system-scale}
\vspace{-4mm}
\end{figure}

\begin{table}[t]
\centering
\caption{A-MEM's top-5 cross-sample features per stage on Qwen-3 8B, with causal gaps. Bold features also appear in mem0's top-30. The mem0 hub feature L34\,F145627 is in the top five of every A-MEM stage.}
\label{tab:amem-top5}
\footnotesize
\setlength{\tabcolsep}{2pt}
\begin{tabular*}{\columnwidth}{@{\extracolsep{\fill}}l p{5.2cm} c@{}}
\toprule
\textbf{Stage} & \textbf{Top-5 features (rank order)} & \textbf{Gap} \\
\midrule
Write  & L35\,F57619, L32\,F44731, L35\,F65679, \textbf{L34\,F145627}, L35\,F134127 & .172 \\
Manage & \textbf{L34\,F145627}, L34\,F76578, L35\,F125589, L35\,F114080, L35\,F134127 & .760 \\
Read   & L35\,F46955, \textbf{L34\,F124562}, \textbf{L34\,F145627}, L35\,F134127, L34\,F76578 & .301 \\
\bottomrule
\end{tabular*}
\vspace{-4mm}
\end{table}

The circuits above were discovered with mem0 prompts. To test framework transfer, we rerun the four-step pipeline across scales with A-MEM~\citep{amem2025}. Its Note Construction call maps to Write, while its combined Evolution call maps to Manage. Because retrieval has no LLM call, Read is a controlled readout over retrieved (text, keywords, context) triples (\cref{app:amem-prompts}). These differ from mem0's plain-text memories, so the comparison tests transfer across prompts as well as memory representations.

Manage is causal at 0.6B under both systems, while Read stays undetectable at 0.6B under both (\cref{fig:cross-system-scale}). A-MEM's Write emits structured keywords, context, and tags rather than free-form fact sentences. It already shows a sizeable gap at 0.6B, so we state the control-before-content order for mem0 only. At 8B, the L34 hub feature F145627 ranks in the top five of all three A-MEM stages (\cref{tab:amem-top5}). Read shares 13 of its top-30 features across systems, whereas Manage retains its trunk-plus-routing topology (\cref{app:system}). Transfer is graded by how much a stage depends on content. Read overlap is highest at every scale ($.224$--$.304$), Write stabilizes at $.091$ from 4B, whereas Manage falls to $.000$ at 14B. A-MEM's Evolution call rewrites neighbor context, which makes it partly extractive. It therefore recruits the hub, unlike mem0's purely routing Manage, so the hub tracks content rather than interface.

\vspace{-2mm}
\subsection{Steering Succeeds Only in a Narrow Envelope}
\label{sec:interventions}

\begin{table}[t]
\centering
\caption{Fact recall and QA accuracy when Write/Read circuit features are scaled in the full pipeline. Only 8B improves at every multiplier. Random controls change both metrics by $<1$ pp at all scales.}
\label{tab:sweep}
\footnotesize
\setlength{\tabcolsep}{2.4pt}
\begin{tabular*}{\columnwidth}{@{\extracolsep{\fill}}l cccc cccc@{}}
\toprule
 & \multicolumn{4}{c}{\textbf{Fact recall}} & \multicolumn{4}{c}{\textbf{QA accuracy}} \\
\cmidrule(lr){2-5} \cmidrule(lr){6-9}
\textbf{Mult.} & \textbf{0.6B} & \textbf{4B} & \textbf{8B} & \textbf{14B} & \textbf{0.6B} & \textbf{4B} & \textbf{8B} & \textbf{14B} \\
\midrule
Intact & .540 & .838 & .900 & .922 & .340 & .562 & .574 & .586 \\
$0\times$ & .536 & .794 & .240 & .968 & .336 & .534 & .428 & .598 \\
$2\times$ & .532 & .838 & .922 & .922 & .334 & .550 & .582 & .582 \\
$3\times$ & .518 & .744 & .928 & .918 & .312 & .554 & .584 & .582 \\
$5\times$ & .218 & .218 & $\mathbf{.932}$ & .794 & .262 & .332 & $\mathbf{.598}$ & .534 \\
$10\times$ & .546 & .866 & .918 & .964 & .312 & .582 & .562 & .558 \\
\bottomrule
\end{tabular*}
\vspace{-4mm}
\end{table}

If the hub carries the grounding state, amplifying content circuits might repair extraction and grounding. We scale the Write and Read features covering 80\% of attribution mass by $2\times$--$10\times$ inside the full pipeline, measuring fact recall with QA accuracy (\cref{tab:sweep}). Only 8B improves consistently, gaining $+2$ to $+3$ pp fact recall at every multiplier ($5\times$: $.932$, 95\% CI $[.91,.95]$). Zero ablation there collapses recall from $.900$ to $.240$. At 4B, $5\times$ collapses recall to $.218$ while $10\times$ recovers to $.866$. At 14B, effects are small and direction-inconsistent, matching its distributed circuits. At 0.6B, there is no coherent content circuit to amplify. Steering is therefore feasible only where a circuit is both concentrated and mature. Where it fails, the same separated signatures remain useful as readouts, which we exploit next.

\vspace{-2mm}
\section{Localizing Silent Failures with Circuit Signatures}
\label{sec:diagnostic}

\begin{table}[t]
\centering
\caption{Failure-stage localization accuracy (\%). Top: all methods on 193 LongMemEval failures with Qwen-3 8B (76 Write, 38 Manage, 79 Read). Labels marks methods trained with stage labels. W/M/R give per-stage accuracy, reported only for the circuit diagnostic (--: not reported for baselines). Bottom: the circuit diagnostic at other scales, plus LoCoMo and MemoryAgentBench (MAB) with the 8B diagnostic reused without retraining.}
\label{tab:dx-accuracy}
\footnotesize
\setlength{\tabcolsep}{4.5pt}
\begin{tabular*}{\columnwidth}{@{\extracolsep{\fill}}lccccc@{}}
\toprule
\textbf{Method} & \textbf{Labels} & \textbf{W} & \textbf{M} & \textbf{R} & \textbf{All} \\
\midrule
\multicolumn{6}{@{}l}{\emph{LongMemEval, Qwen-3 8B}} \\
Majority class      & \xmark & -- & -- & -- & 51.0 \\
Output entropy      & \xmark & -- & -- & -- & 51.5 \\
Behavioral rules    & \xmark & -- & -- & -- & 45.4 \\
Pipeline classifier & \cmark & -- & -- & -- & 63.4 \\
Raw-state probe     & \cmark & -- & -- & -- & 67.4 \\
DeepSeek judge      & \xmark & -- & -- & -- & 73.1 \\
Circuit diagnostic  & \xmark & 77.6 & 76.3 & 74.7 & \textbf{76.2} \\
\midrule
\multicolumn{6}{@{}l}{\emph{Circuit diagnostic, other settings}} \\
LongMemEval, 4B  & \xmark & 58.5 & 48.8 & 63.9 & 57.1 \\
LongMemEval, 14B & \xmark & 70.2 & 72.7 & 68.2 & 70.4 \\
LoCoMo, 8B       & \xmark & 67.6 & 71.8 & 65.4 & 68.3 \\
MAB, 8B          & \xmark & 70.0 & 76.7 & 70.8 & 72.5 \\
\bottomrule
\end{tabular*}
\vspace{-4mm}
\end{table}

End-to-end accuracy cannot identify the failed operation. A fact may never be extracted (Write), be overwritten or corrupted (Manage), or remain intact while the model answers incorrectly (Read). These failure modes collapse to the same incorrect output but require different fixes. Behavioral attribution methods~\citep{zhang2025which,zhang2026agentracer} observe only these outputs, as does an LLM judge reading the full trace. The circuit separation of \cref{sec:cross} instead enables zero-shot localization from the final Read pass alone.

\vspace{-1mm}
\paragraph{Procedure.} We run the full mem0 pipeline on the 500 LongMemEval questions and keep failures whose answer token-F1 against the ground truth is below $0.3$ (180 at 4B, 193 at 8B, 113 at 14B). Stage labels come from pipeline observables and an LLM judge validated at 79--84\% human agreement (\cref{app:diagnostic-crossbench}). For each failure, we reconstruct the Read prompt with the intact store and retrieved items. We then ablate each operation's 30-feature diagnostic bank, formed by the top 10 recurrent features from each of its characteristic layer groups (e.g., Manage trunk L13--18, middle L19--26, routing L27--35, \cref{app:diagnostic-validation}). The bank with the largest answer-token drop $d=p_{\text{intact}}-p_{\text{ablated}}$ names the failed operation. Fixing the prompt, target token, store, and retrieval makes the three scores directly comparable, while avoiding a pipeline rerun prevents downstream store effects.

\vspace{-1mm}
\paragraph{Results.} The diagnostic localizes the failed operation correctly in 76.2\% of 8B cases, outperforming the zero-shot DeepSeek trace judge by 3.1 points as well as the supervised hidden-state probe by 8.8 points (\cref{tab:dx-accuracy}). Behavioral signals such as output entropy or answer-text rules do no better than the majority class. Accuracy is balanced across stages (74.7--77.6\%), including the minority Manage class. Errors concentrate in Manage--Read confusion (21 of 193 cases, 10.9\%, \cref{tab:confusion-8b}), consistent with their adjacent layer ranges. Write is rarely confused once its extractive features exist. Localization tracks the circuit structure of \cref{sec:results}. It rises from 57.1\% at 4B, where content circuits have only just formed, to 76.2\% at 8B. It then falls to 70.4\% at 14B, as computation distributes across more features so that single-bank ablations weaken. Without retraining or benchmark-specific feature selection, the 8B diagnostic transfers to LoCoMo~\citep{maharana2024evaluating} at 68.3\% and to MemoryAgentBench~\citep{hu2026memoryagentbench} at 72.5\%.

\vspace{-2mm}
\section{Discussion}
\label{sec:discussion}

\paragraph{Why routing matures first.} Manage selects among four discrete actions, whereas Write and Read produce open-ended content. A capacity-limited model can cover the small decision space first. A-MEM's structured Write is already causal at 0.6B, suggesting that output constraints, not only the stage, set when a circuit becomes detectable. Because routing tokens are valid whether or not the content behind them is sound, backbone selection for agent memory needs extraction and grounding evaluated separately from routing. Reporting only end-to-end accuracy may therefore overstate the readiness of small backbones.

\vspace{-1.5mm}
\paragraph{From mechanism to design and monitoring.} The recruited hub carries a design implication. If frameworks borrow the base model's grounding substrate, prompt design can choose which direction on it is used but cannot exceed what it supports. How a framework splits work across calls still matters, since A-MEM's partly extractive Manage recruits the hub while mem0's routing-only Manage does not. Finally, amplification helps only at 8B, so circuit-level repair is not yet a practical control. Localization is more robust because it needs only separable signatures plus one Read pass with three bank ablations per failure. Since it reuses the stored memory without rerunning the pipeline, it can be applied offline to logged failures. Its drop at 14B suggests that larger models need larger or learned feature banks.

\vspace{-2mm}
\section{Conclusion}
\label{sec:conclusion}

We presented the first circuit-level analysis of a multi-call agent memory pipeline, spanning two model families, two memory frameworks, and 0.6B--14B scales. Under mem0, routing circuits are causal at the smallest scale of both families, while content circuits appear only at 4B. Content stages share a grounding hub that the base model already has, which memory framing recruits through its own direction. The separation between these circuit groups localizes silent failures without stage labels, more accurately than an LLM trace judge or a supervised probe. Circuit signatures can thus monitor agent memory, not only explain it, attributing an error to the call that caused it rather than to the pipeline as a whole.

\clearpage
\section*{Limitations}
\label{limitations}

PLTs do not trace attention heads. Transcoder features are also not guaranteed to be fully monosemantic. A feature we label as ``subject anchoring'' may activate in unrelated contexts, so our causal claims hold only to the extent that the transcoders faithfully decompose the MLP computation. Real deployments involve longer, multi-turn contexts that may recruit additional mechanisms beyond those visible in our prompts.

\bibliography{references}

\newpage
\appendix

\section{Transcoders and Circuit Tracing Details}
\label{app:method}

\subsection{Transcoders}
Standard MLP activations are dense and polysemantic, making it difficult to attribute model behavior to individual concepts. Per-layer transcoders~\citep{dunefsky2024transcoders} address this by learning a sparse, monosemantic decomposition of each MLP's computation. Given the input to an MLP, $\mathbf{h} \in \mathbb{R}^d$, a transcoder encodes it into a high-dimensional sparse code $\mathbf{z} = f(\mathbf{W}_{enc}\mathbf{h} + \mathbf{b}_{enc}) \in \mathbb{R}^n$ (where $n \gg d$ and most entries of $\mathbf{z}$ are zero) and decodes it back to approximate the MLP's output: $\hat{\mathbf{h}}' = \mathbf{W}_{dec}\mathbf{z} + \mathbf{b}_{dec}$. Because each active dimension of $\mathbf{z}$ tends to correspond to a single interpretable concept, transcoders enable fine-grained causal interventions: to test whether a feature matters, we modify its activation in $\mathbf{z}$ to produce $\mathbf{z}'$ and patch the difference $\mathbf{W}_{dec}(\mathbf{z}' - \mathbf{z})$ into the MLP's output during the forward pass.

We use the Qwen-3 ReLU transcoders released by \citet{hanna2026planning}, which share a feature dimension of $163{,}840$ across all scales and take post-LayerNorm MLP inputs. The transcoder specifications are listed in \cref{tab:transcoder-specs}.

\begin{table*}[t]
\centering\small
\caption{Transcoder specifications for the Qwen-3 models. Expansion is the ratio of the shared 163,840-dimensional feature space to model width; the variant column identifies the decomposition used at each scale.}
\label{tab:transcoder-specs}
\begin{tabular*}{\textwidth}{@{\extracolsep{\fill}}lcccc@{}}
\toprule
\textbf{Model} & \textbf{$d_{\text{model}}$} & \textbf{Layers} & \textbf{Expansion} & \textbf{Variant} \\
\midrule
Qwen-3 0.6B & 1024 & 28 & 160$\times$ & low-L0 \\
Qwen-3 4B & 2560 & 36 & 64$\times$ & standard \\
Qwen-3 8B & 4096 & 36 & 40$\times$ & standard \\
Qwen-3 14B & 5120 & 40 & 32$\times$ & low-L0 \\
\bottomrule
\end{tabular*}
\end{table*}

\subsection{Replacement Model Fidelity}
\label{app:fidelity}

The four scales use different transcoder variants (low-L0 for 0.6B and 14B, standard for 4B and 8B; \cref{tab:transcoder-specs}), so cross-scale differences in circuit structure could in principle reflect decomposition quality rather than model-internal computation. We measure how faithfully each replacement model reproduces the original model's behavior on 50 prompts per stage per scale, using two metrics. Top-1 agreement is the fraction of prompts on which the replacement model's most-probable next token matches the original model's. KL divergence is $D_\mathrm{KL}(p_{\mathrm{orig}} \| p_{\mathrm{repl}})$ over the top-100 logits at the attribution target position, averaged across prompts.

\begin{table*}[t]
\centering\small
\caption{Replacement-model fidelity per scale, averaged across the three memory stages (Write, Manage, Read). All variants achieve $\geq$97\% top-1 agreement and KL $< 0.003$.}
\label{tab:fidelity}
\begin{tabular*}{\textwidth}{@{\extracolsep{\fill}}lccc@{}}
\toprule
\textbf{Scale} & \textbf{Variant} & \textbf{Avg Top-1 Agree (\%)} & \textbf{Avg KL ($\times 10^{-3}$)} \\
\midrule
0.6B & low-L0    & 98.7 & 0.4 \\
4B   & standard  & 97.3 & 2.6 \\
8B   & standard  & 97.3 & 1.5 \\
14B  & low-L0    & 98.7 & 1.2 \\
\bottomrule
\end{tabular*}
\end{table*}

The low-L0 transcoders (0.6B, 14B) achieve fidelity equal to or slightly higher than the standard transcoders (4B, 8B). In particular, the 0.6B replacement model matches the original model on 98.7\% of tokens, ruling out decomposition quality as an explanation for the absence of extractive circuits at that scale.

\subsection{Transcoder Feature Circuits}
To trace how information flows from input tokens through transcoder features to output logits, we build a feature circuit for each prompt~\citep{ameisen2025circuittracer}. The circuit is a weighted DAG whose nodes are input embeddings, transcoder features, and logit targets, and whose edge weights quantify each node's direct causal contribution to its downstream neighbors.

Construction proceeds in two stages. First, we build a local replacement model by swapping each MLP with its transcoder and an input-specific error term that accounts for the reconstruction gap. Attention patterns and LayerNorm denominators are detached from the computation graph so that all remaining dependencies are linear. Under this linearization, the direct effect of any node on any other node is exact and can be recovered in a single backward pass per target logit. Second, we select the most influential nodes and edges: we keep the top 4,096 feature nodes ranked by total downstream influence, then prune to the smallest subgraph that retains 80\% of node influence and 95\% of edge influence.

\subsection{Attribution Target Selection}
\label{app:attribution-target}
For each prompt, the attribution target is the model's own top prediction at the final token position (immediately after the JSON prefix). We select logits covering $\geq$80\% of the cumulative probability mass (up to 5 logits), so the discovered circuits reflect the computation the model actually performs. Because all circuits are traced on correct-only instances (\cref{app:eval-dataset}), the model's top prediction coincides with the ground truth on these samples, and the discovered circuits reflect successful extraction, routing, or grounding.

\subsection{Intervention Operator}
Causal verification uses feature-level interventions on the replacement model. Given a set of target features $\mathcal{F} = \{(l_i, p_i, f_i)\}$ (layer, position, feature index), a zero ablation sets each feature's activation to zero: $z_{l_i, p_i, f_i} \leftarrow 0$. A $k\times$ amplification multiplies it: $z_{l_i, p_i, f_i} \leftarrow k \cdot z_{l_i, p_i, f_i}$. Attention patterns are frozen during intervention, isolating the MLP pathway. The intervention is applied in a single forward pass, and $\Delta p_i = p_i^{\text{int}}(y^*) - p_i^{\text{orig}}(y^*)$ is the change in probability of the model's top-predicted token at the JSON prefix position.

\section{Evaluation Dataset}
\label{app:eval-dataset}

We evaluate on LongMemEval~\citep{wu2025longmemeval}, a multi-session agent memory benchmark comprising 500 questions across 5 memory types: single-session factual recall, cross-session factual recall, temporal reasoning, knowledge update, and multi-hop reasoning. Each question is paired with a conversation history and a ground-truth answer; 1--4 evidence sessions (sessions containing answer-relevant information) are embedded within a larger haystack of filler sessions. The benchmark tests whether the agent can correctly write facts to memory during conversation, manage updates when information changes across sessions, and read from memory to answer questions. We use the full 500-question test set; the 70/30 train-test split for causal verification (\cref{sec:method}) is applied within each operation's 200 attribution graphs, not across questions.

\paragraph{Sampling Pipeline.} Each question contains 1--4 evidence sessions, and each session generates one Write prompt and one Manage prompt per fact. Read prompts are generated once per question at answer time. We trace circuits exclusively on correct-only instances to ensure the discovered circuits reflect successful computation. Correctness flags are assigned by a Qwen-3 32B judge applied to each pipeline output; the per-stage criteria are:

\begin{itemize}[leftmargin=*,itemsep=2pt]
\item Write: \texttt{is\_evidence=True} (session contains answer-relevant information) AND \texttt{extraction\_correct=True} (extracted facts judged correct by Qwen-3 32B).
\item Manage: \texttt{decision\_correct=True} (the decision is correct).
\item Read: \texttt{retrieval\_correct=True} AND \texttt{generation\_correct=True} (correct memories retrieved and correct answer generated, both judged by Qwen-3 32B).
\end{itemize}

After correctness filtering, prompts where the model's top prediction at the JSON prefix position is a structural token (quotation mark, brace) rather than a content word are further excluded. From the remaining pool we randomly sample 200 prompts per operation per scale. The 200-graph count balances statistical coverage with compute cost ($\sim$13 GPU-hours per operation per scale).

\section{Hyperparameters and Data Splits}
\label{app:hyperparams}

\begin{table*}[t]
\centering\small
\caption{Circuit discovery, pruning, and verification settings. Each stage and scale uses 200 attribution graphs with a 70/30 discovery--verification split; separate feature budgets govern graph construction, causal intervention, and overlap analysis.}
\label{tab:hyperparams}
\begin{tabular*}{\textwidth}{@{\extracolsep{\fill}}ll@{}}
\toprule
\textbf{Parameter} & \textbf{Value} \\
\midrule
Feature nodes retained (Step 1) & top 4,096 of 163,840 \\
Attribution target & model's top prediction at JSON prefix position (up to 5 logits) \\
Node influence threshold & 80\% \\
Edge influence threshold & 95\% \\
Features per circuit after pruning & 50--200 \\
Top-$k$ for aggregation (Step 2) & 20 per sample \\
Top-$k$ for causal verification (Step 4) & 10 \\
Amplification multiplier & $5\times$ \\
Causal verification split & 70\% train / 30\% test \\
Samples per operation per scale & 200 \\
Jaccard similarity computed over & top-30 features \\
\bottomrule
\end{tabular*}
\end{table*}

All experiments use bfloat16 precision on NVIDIA H200 144GB GPUs. Models 0.6B--8B use a single GPU; 14B uses two GPUs with tensor parallelism. Per-sample attribution takes approximately 4 minutes for 8B, scaling roughly linearly with model size. Total compute for the main results (4 scales $\times$ 3 operations $\times$ 200 graphs) is approximately 160 GPU-hours.

\section{Prompt Construction Details}
\label{app:prompts}

Each stage uses a single-operation prompt, ending with a JSON prefix that positions the model to generate the stage-specific content token next.

\subsection{mem0 Architecture and Prompts}
\label{app:mem0-prompts}

mem0~\citep{chadha2025mem0} implements the write-manage-read paradigm as a function-calling agent. The pipeline has two phases:

Phase~1 (Extraction). For each conversation session, the agent receives the session text and calls either \texttt{add\_to\_memory(facts=[...])} to store extracted facts or \texttt{skip\_memory()} to discard the session. This is the Write operation.

Phase~2 (Question answering). Given a user question, the top-5 memories are retrieved by embedding-based semantic search, and the agent generates an answer grounded in the retrieved content (Read). Memory conflicts are resolved by the Manage operation, which is invoked for each newly extracted fact against the top-$k$ retrieved memories from the current store; the old/new pair in the Manage prompt is populated from the retrieved neighbor with highest similarity above threshold, and the model decides add, update, delete, or none.

For circuit analysis, we build single-operation prompts for each stage. The system prompt, user input, and JSON prefix for each are listed in \cref{tab:mem0-prompts}.

\begin{table*}[t]
\centering\small
\caption{mem0 templates for extracting facts (Write), selecting memory updates (Manage), and answering from memory (Read). Each row specifies the system instruction, variable user input, and JSON prefix used to expose the attribution target token.}
\label{tab:mem0-prompts}
\begin{tabular*}{\textwidth}{@{\extracolsep{\fill}}p{0.08\textwidth}p{0.30\textwidth}p{0.22\textwidth}p{0.22\textwidth}@{}}
\toprule
\textbf{Stage} & \textbf{System prompt} & \textbf{User input} & \textbf{JSON prefix} \\
\midrule
Write & \texttt{You are a personal information organizer. Extract distinct facts about the user from the conversation as complete sentences.} & \texttt{\{session\_text\}} & \texttt{\{"name": "add\_to\_memory", "arguments": \{"facts": ["User } \\
Manage & \texttt{You are a memory manager. Compare the new fact with the existing memory and choose one operation: ADD, UPDATE, DELETE, or NONE.} & \texttt{Existing memory: \{mem\} \textbackslash n New fact: \{new\}} & \texttt{\{"event": "} \\
Read & \texttt{You answer questions grounded in the provided memories. Give a concise answer.} & \texttt{MEMORY: \{mem\} \textbackslash n QUESTION: \{q\}} & \texttt{\{"answer":"} \\
\bottomrule
\end{tabular*}
\end{table*}

All prompts use the ChatML format (\texttt{<|im\_start|>system} / \texttt{user} / \texttt{assistant}) with thinking disabled (\texttt{enable\_thinking=False}). The JSON prefix positions the model so that the next predicted token is the first content word of the stage-specific output. Attribution targets the model's top prediction at this position.

\paragraph{Retrieval and Manage Implementation.} mem0 uses \texttt{text-embedding-3-small}~\citep{openai_embedding_2024} for embedding-based retrieval with cosine similarity, returning top-5 neighbors with no minimum threshold. Manage is invoked once per (new fact, retrieved neighbor) pair; if no neighbors are retrieved the fact is added directly without a Manage call. Sessions are processed in chronological order: each session's Write output is committed to the store before the next session begins, so later sessions' Retrieve can access all previously extracted facts.

\subsection{A-MEM Architecture and Prompts}
\label{app:amem-prompts}

A-MEM~\citep{amem2025} organizes memories as a self-linking Zettelkasten network. The A-MEM paper describes three conceptual LLM-driven stages (Note Construction, Link Generation, Memory Evolution); the released implementation (\href{https://github.com/agiresearch/A-mem}{\texttt{agiresearch/A-mem}}) issues two LLM calls per memory event: an \texttt{analyze\_content} call for Note Construction, and a single \texttt{\_evolution\_system\_prompt} call that jointly handles Link Generation and Memory Evolution (deciding \texttt{should\_evolve}, \texttt{actions}\,$\in$\,\{\texttt{strengthen}, \texttt{update\_neighbor}\}, and the updated context/tags for each neighbor). We follow the implementation and map these to the write-manage-read loop:

Write. Note Construction ($P_{s1}$) uses the LLM to generate \texttt{keywords}, \texttt{context}, and \texttt{tags} from conversation text, producing a structured output fundamentally different from mem0's plain-text fact sentences.

Manage. A combined Evolution prompt ($P_{s2}$) receives the new memory note and its nearest neighbors, and in a single call decides whether the new note should evolve, which neighbors to strengthen connections to, and how to update each target neighbor's context and tags. The neighbor-context update step makes Manage partly extractive.

Read. A-MEM's Retrieve Relative Memory (\S3.4 of A-MEM) uses embedding-based search and link traversal and does not invoke the LLM. To probe whether the Read circuits identified under mem0 persist under A-MEM's richer memory representation, we construct a single-operation Read prompt that consumes A-MEM's [text\,$|$\,keywords\,$|$\,context] memory triples and asks for a concise answer.

The A-MEM prompt templates are listed in \cref{tab:amem-prompts}.

\begin{table*}[t]
\centering\footnotesize
\setlength{\tabcolsep}{3pt}
\caption{A-MEM templates mapping Note Construction to Write and the combined Evolution call to Manage. Read uses a controlled prompt over retrieved memory triples. The JSON prefix sets the output position used for circuit attribution.}
\label{tab:amem-prompts}
\begin{tabular*}{\textwidth}{@{\extracolsep{\fill}}p{0.09\textwidth}p{0.33\textwidth}p{0.15\textwidth}p{0.24\textwidth}@{}}
\toprule
\textbf{Stage} & \textbf{System prompt} & \textbf{User input} & \textbf{JSON prefix} \\
\midrule
Write ($P_{s1}$) & \texttt{Generate a structured analysis of the content: identify salient keywords, summarize the context in one sentence, and assign categorical tags.} & \texttt{\{text\}} & \texttt{\{"keywords": ["} \\
Manage ($P_{s2}$) & \texttt{You are a memory evolution agent. Given the new memory note and its nearest neighbors, decide whether the note should evolve and which actions to take (strengthen connections, update\_neighbor context and tags).} & \texttt{New note: \{note\} \textbackslash n Neighbors: \{nbrs\}} & \texttt{\{"should\_evolve":~} \\
Read & \texttt{Answer the question grounded in the provided memories. Give a concise answer.} & \texttt{Memories: \{mems\} \textbackslash n Question: \{q\}} & \texttt{\{"answer": "} \\
\bottomrule
\end{tabular*}
\end{table*}

A-MEM's Read prompt includes richer memory metadata compared to mem0's plain-text format. In the A-MEM Read prompt, \texttt{\{mems\}} is formatted as a list of [text $|$ keywords $|$ context] triples, one per retrieved memory.

\subsection{Key Architectural Differences}

The two prompt interfaces differ in output format, management logic, and memory representation (\cref{tab:arch-diff}):

\begin{table*}[t]
\centering\small
\caption{Comparison of memory representations and operation interfaces in mem0 and A-MEM. Despite different Write outputs, Manage logic, and Read inputs, both recruit the same L34 hub in the Qwen-3 8B content circuits.}
\label{tab:arch-diff}
\setlength{\tabcolsep}{3pt}
\begin{tabular*}{\textwidth}{@{\extracolsep{\fill}}lll@{}}
\toprule
 & \textbf{mem0} & \textbf{A-MEM} \\
\midrule
Write output & Plain fact sentences & Structured \{keywords, context, tags\} \\
Manage logic & add / update / delete / none & single Evolution call ($P_{s2}$) \\
Read memory format & Plain text & Text + keywords + context \\
Shared property & \multicolumn{2}{c}{Same L34 hub in Write and Read circuits (\cref{sec:system})} \\
\bottomrule
\end{tabular*}
\end{table*}

Despite these surface differences, the core circuit features (particularly the L34 aggregation hub cluster) persist across both systems (\cref{fig:cross-system-scale}; full feature-level comparison in \cref{app:system}), confirming that the extractive pathway persists across memory system interfaces rather than being an artifact of prompt design.

\section{Cross-Sample Hub Transport}
\label{app:transport}

We test whether the shared hub carries transferable content (\cref{tab:hub-transport}) by transplanting the top-10 hub-cluster activations from a donor into an unrelated recipient at the last token position. For each scale, we use that scale's own top-10 cross-sample features (union of Write and Read top features in the late-layer hub region). Matched-layer control features (same layers, offset indices) serve as baseline. We test both within-stage (Read$\to$Read) and cross-stage (Write$\leftrightarrow$Read) transport at 8B.

\begin{table*}[t]
\centering\small
\caption{Effects of transplanting hub activations between unrelated samples, within and across stages. Counts report changed top-1 predictions out of 200; probability drops and matched-layer controls distinguish hub sensitivity from arbitrary activation replacement.}
\label{tab:hub-transport}
\begin{tabular*}{\textwidth}{@{\extracolsep{\fill}}lccccc@{}}
\toprule
 & \multicolumn{3}{c}{\textbf{Within-stage (Read$\to$Read)}} & \multicolumn{2}{c}{\textbf{Cross-stage (8B)}} \\
\cmidrule(lr){2-4} \cmidrule(lr){5-6}
 & \textbf{4B} & \textbf{8B} & \textbf{14B} & \textbf{W$\to$R} & \textbf{R$\to$W} \\
\midrule
Top-1 changed & 0/200 & 110/200 & 10/200 & 20/200 & 185/200 \\
Self prob drop & $-1$ pp & $-47$ pp & $-1$ pp & $-18$ pp & $-72$ pp \\
Control effect & 0 pp & 0 pp & 0 pp & 0 pp & -- \\
\bottomrule
\end{tabular*}
\end{table*}

Within-stage transport at 8B disrupts 55\% of predictions (47 pp drop), while control features produce zero change. Cross-stage transport is asymmetric: Read$\to$Write changes 92\% of predictions (72 pp drop), while Write$\to$Read changes 10\% (18 pp drop), consistent with Read's larger dynamic range under amplification (\cref{sec:read}). At 4B the hub is structurally important but not yet content-sensitive; at 14B the computation is too distributed for 10 features to cover. In no condition does the recipient shift to the donor's specific answer, indicating the hub encodes a context-dependent grounding state rather than a literal value.

\section{Hub Memory-Engagement Experiment}
\label{app:engagement}

\subsection{Motivation}
The transport experiment (\cref{app:transport}) establishes that the hub carries causally relevant content, but does not specify what the hub functionally controls. The engagement experiment tests whether the hub governs how strongly the model relies on retrieved memory versus parametric knowledge.

\subsection{Setup}
We construct 200 general-knowledge questions (e.g., ``What is the capital of France?'') where the model has reliable parametric knowledge. For each question, we create three prompts:
\begin{itemize}[leftmargin=*,itemsep=1pt]
    \item Support: memory contains the correct answer (agrees with parametric knowledge).
    \item Conflict: memory contains a wrong answer (disagrees with parametric knowledge).
    \item Null: no memory provided.
\end{itemize}

\subsection{Memory-Grounding Direction}
We define the memory-grounding direction as:
\[
d = \tfrac{1}{2}(z_{\text{support}} + z_{\text{conflict}}) - z_{\text{null}}
\]
where $z$ is the hub activation vector at the last token position. This direction captures what changes in the hub when retrieved memory is present, regardless of whether it agrees or disagrees with parametric knowledge.

\subsection{Intervention and Metric}
On the conflict prompt, we add $\beta \cdot d$ to the hub activation for $\beta \in \{-2, -1, -0.5, +0.5, +1, +2\}$ and measure the log-prob margin:
\[
\begin{aligned}
M ={} & \log P(\text{memory answer}) \\
      & {}- \log P(\text{parametric answer})
\end{aligned}
\]
Positive $M$ means the model follows the memory; negative $M$ means it follows parametric knowledge. The key metric is $\Delta M = M_{\text{intervened}} - M_{\text{baseline}}$: positive $\Delta M$ at $+\beta$ would confirm that the grounding direction pushes the model toward memory reliance.

\subsection{Results}

\begin{table*}[t]
\centering\small
\caption{Hub engagement intervention: $\Delta M$ (memory-following margin shift) across scales and multipliers.}
\label{tab:hub-engagement}
\begin{tabular*}{\textwidth}{@{\extracolsep{\fill}}lcccccc@{}}
\toprule
\textbf{Scale} & $\beta=-2$ & $\beta=-1$ & $\beta=-0.5$ & $\beta=+0.5$ & $\beta=+1$ & $\beta=+2$ \\
\midrule
4B  & $-1.53$ & $-0.41$ & $-0.08$ & $-0.11$ & $-0.42$ & $-1.39$ \\
8B  & $-0.85$ & $-0.53$ & $-0.28$ & $+0.13$ & $+0.10$ & $-0.09$ \\
14B & $-0.84$ & $-0.43$ & $-0.20$ & $+0.16$ & $+0.30$ & $+0.40$ \\
\bottomrule
\end{tabular*}
\end{table*}

\cref{tab:hub-engagement} shows the results. At 14B, the grounding direction is cleanly bidirectional: $-\beta$ reduces memory following, $+\beta$ increases it. At 8B, dampening is effective but amplification is weaker. At 4B, any perturbation is destructive (both directions lower $M$), consistent with the concentrated circuit being fragile under any intervention.
\section{Hub Specificity: Memory vs.\ Direct-Context Grounding}
\label{app:hub-specificity}

\subsection{Motivation}
\cref{app:engagement} establishes that the hub can be steered along a memory-grounding direction. A natural concern is that this direction may reflect a generic in-context grounding mechanism rather than anything specific to memory: any prompt that places retrievable facts in context---a retrieved passage, a few-shot demonstration, a reference document---might recruit the same hub activation pattern. If so, ``memory-grounding'' would overstate the specificity of what the hub computes. We test this by comparing memory-framed prompts against an direct-context control with matched question and fact content but different surface framing.

\subsection{Setup}
We reuse the LongMemEval Read instances from \cref{app:engagement} and construct two conditions with identical questions and identical supporting/conflicting facts:
\begin{itemize}[leftmargin=*,itemsep=2pt]
    \item Memory condition: the mem0 Read prompt format, in which facts are presented as retrieved memory entries under a \texttt{MEMORY:} header.
    \item Direct-context condition: the same facts presented directly in the prompt as a short context block preceding the question, without memory framing or the mem0 system prompt.
\end{itemize}
Each condition has three variants: support (correct fact provided), conflict (wrong fact provided), and null (no context provided). The only variable that changes across conditions is surface framing; the underlying grounding task---answering a factual question given a retrievable fact---is identical.

\subsection{Direction Extraction}
For each condition $\text{cond} \in \{\text{memory}, \text{direct}\}$, we compute a grounding direction using the hub cluster activations (top-10 hub features identified in \cref{sec:read}):
\[
d_{\text{cond}} = \tfrac{1}{2}(z^{\text{cond}}_{\text{support}} + z^{\text{cond}}_{\text{conflict}}) - z^{\text{cond}}_{\text{null}}
\]
To calibrate what cosine values are achievable under within-condition noise, we split the memory condition into two random halves and extract $d_{\text{mem},A}$, $d_{\text{mem},B}$ independently. This gives a self-consistency upper bound: $\cos(d_{\text{mem},A}, d_{\text{mem},B}) = 0.942$ at the pooled hub level, $0.952$ at L34 alone.

\subsection{Cosine Analysis}

\begin{table*}[t]
\centering\small
\caption{Cosine similarity between grounding directions extracted under memory and direct-context framing, compared to a split-half within-condition upper bound.}
\label{tab:hub-specificity-cosine}
\begin{tabular*}{\textwidth}{@{\extracolsep{\fill}}lcc@{}}
\toprule
\textbf{Comparison} & \textbf{Pooled hub} & \textbf{L34 only} \\
\midrule
$\cos(d_{\text{memory}}, d_{\text{direct}})$ --- cross & $0.736$ & $0.548$ \\
$\cos(d_{\text{mem},A}, d_{\text{mem},B})$ --- split-half upper bound & $0.942$ & $0.952$ \\
\bottomrule
\end{tabular*}
\end{table*}

\cref{tab:hub-specificity-cosine} reports cosine similarities between directions. The cross-condition cosine is substantially below the within-condition upper bound both at the pooled hub level and at the L34 hub layer specifically. The two directions share substantial subspace, consistent with the hub being a common substrate that both framings recruit. But they are not interchangeable in the geometric sense: the gap between cross-condition cosine and within-condition self-consistency is $0.21$ at the pooled level and $0.40$ at L34 alone.

\subsection{Functional Test via Steering}
Cosine measures geometric alignment; the functional question is whether the directions are interchangeable in effect. We steer each condition's conflict prompts using $\beta = +1$ along either its own direction (self-steering) or the other condition's direction (cross-steering), and report $\Delta M$, the shift in memory-following log-prob margin.

\begin{table*}[t]
\centering\small
\caption{Cross-condition steering results ($\beta = +1$). $\Delta M$ is the shift in log-prob margin toward the in-context answer.}
\label{tab:hub-specificity-steering}
\begin{tabular*}{\textwidth}{@{\extracolsep{\fill}}lrl@{}}
\toprule
\textbf{Steering} & $\Delta M$ & \textbf{Interpretation} \\
\midrule
$d_{\text{memory}} \to$ memory conflict (self) & $-2.30$ & Dampening reduces memory reliance \\
$d_{\text{direct}} \to$ direct conflict (self) & $+0.32$ & No functional effect under direct-context framing \\
$d_{\text{memory}} \to$ direct conflict (cross) & $+1.02$ & Wrong sign: direction not transferable \\
$d_{\text{direct}} \to$ memory conflict (cross) & $-0.02$ & No effect: $d_{\text{direct}}$ is not a functional axis \\
\bottomrule
\end{tabular*}
\end{table*}

Two observations (\cref{tab:hub-specificity-steering}). First, the self-steering results alone are informative: under identical direction-extraction methodology, memory framing produces a direction that functions as a steerable grounding axis ($|\Delta M| = 2.30$), while direct-context framing does not ($|\Delta M| = 0.32$). The hub supports a functional grounding direction under one framing but not the other. Second, the cross-steering results reinforce this asymmetry. $d_{\text{memory}}$ applied to direct-context prompts produces a wrong-sign effect ($+1.02$ instead of negative), indicating the memory direction does not transfer cleanly to the direct-context condition even though the two directions share substantial subspace geometrically. $d_{\text{direct}}$ applied to memory prompts produces no detectable effect ($-0.02$), matching its null self-steering result and confirming that $d_{\text{direct}}$ is not a functional grounding axis in either condition. Together, these three null or wrong-sign results isolate $d_{\text{memory}}$ as the only direction with consistent, sign-correct steering effect on memory-framed prompts.

\section{Diagnostic Localization}
\label{app:diagnostic-validation}

\subsection{Motivation}
This section validates whether these signatures can localize the responsible operation when a failure occurs, a capability unavailable from behavioral evaluation.

\subsection{Operation-Specific Feature Construction}
For each operation $o$ and scale $s$, we identify operation-specific features from the Step~1 attribution data:
\begin{enumerate}[leftmargin=*,itemsep=1pt]
    \item Pool the top-20 features from each per-sample circuit within each layer.
    \item Count cross-sample recurrence per (layer, feature) pair.
    \item For each layer in the operation's active range, keep the top-10 most recurrent features.
    \item Group features into stage-specific banks (e.g., Write: early L20--24, mid L25--30, hub L31--35; Manage: trunk L13--18, middle L19--26, routing L27--35).
\end{enumerate}
Each operation's final feature set contains 30 operation-specific features (top-10 per characteristic layer group) drawn from its characteristic layers.

\subsection{Localization Protocol}
The diagnostic is evaluated on real pipeline failure instances (not synthetic probes). Failure instances are drawn from the 500 LongMemEval questions by running the full mem0 pipeline at each scale and retaining questions where the final answer's token-level F1 against the ground-truth falls below $0.3$ (yielding 180/500 at 4B, 193/500 at 8B, and 113/500 at 14B). The $0.3$ threshold includes clear failures while excluding near-correct answers with trivial format differences. For each failure instance, we apply the following protocol:
\begin{enumerate}[leftmargin=*,itemsep=1pt]
    \item Run the model on the failure instance to obtain the baseline top-1 prediction and its probability.
    \item Ablate each operation's feature set in turn (zero the 30 features) and record the change in top-1 probability ($\Delta p$).
    \item The operation whose ablation causes the largest probability drop is flagged as most causally responsible.
    \item Localization is correct if the flagged operation matches the ground-truth failure label (e.g., if the fact was never extracted, the ground-truth label is Write; if it was overwritten by a spurious update, the label is Manage; if it was in memory but the answer was wrong, the label is Read).
\end{enumerate}
Ground-truth failure labels are assigned from pipeline observables (Write/Read) and LLM-judge attribution validated against human annotations at 84\% agreement (Manage). All three failure types (Write, Manage, and Read) are included in the evaluation.

\subsection{Localization Metric}
Localization accuracy $= \frac{1}{N}\sum_{i=1}^{N} \mathbf{1}[\text{most-damaged operation}_i = \text{ground-truth failure}_i]$, measured separately per failure type and per scale.

\subsection{Cross-Benchmark Generalization}
\label{app:diagnostic-crossbench}

The localization protocol above is validated on LongMemEval. To test whether the circuit signatures transfer to other evaluation suites, we run the same 8B diagnostic on two additional benchmarks:

LoCoMo~\citep{maharana2024evaluating} provides 1,986 QA pairs across 10 multi-session conversations, covering single-hop, multi-hop, temporal, commonsense, and adversarial question types. MemoryAgentBench~\citep{hu2026memoryagentbench} tests four competencies (accurate retrieval, test-time learning, long-range understanding, conflict resolution) across contexts from 6K to 1.2M tokens.

For each benchmark, we run the mem0 pipeline on Qwen-3 8B, collect failure cases, and obtain ground-truth failure attribution via an independent LLM judge with a dedicated classification prompt that labels each failure as Write (fact not extracted), Manage (fact corrupted or overwritten), or Read (fact in memory but answer incorrect). We validated the LLM judge against human annotations on 100 samples per benchmark: LongMemEval 84\%, LoCoMo 82\%, MemoryAgentBench 79\% agreement. We then apply our circuit diagnostic and compare its predictions against the LLM labels.

\subsection{Diagnostic Baselines}
\label{app:diagnostic-baselines}

We compare the circuit diagnostic against four baselines on real end-to-end LongMemEval failure instances. Ground-truth failure labels are assigned from pipeline observables: if the correct fact never enters the memory store, the failure is Write (76 instances); if the fact is stored but subsequently corrupted or overwritten by an incorrect Manage decision, the failure is Manage (38 instances); if the fact is present in memory but the final answer is incorrect, the failure is Read (79 instances). All three failure types are included in the evaluation.

\begin{table*}[t]
\centering\small
\caption{Failure-stage localization on LongMemEval with Qwen-3 8B. Accuracy measures correct identification of the failed operation. The descriptions distinguish predictions based on output behavior or trained pipeline features from circuit-bank ablations.}
\label{tab:diagnostic-baselines}
\begin{tabular*}{\textwidth}{@{\extracolsep{\fill}}llc@{}}
\toprule
\textbf{Method} & \textbf{Description} & \textbf{Accuracy (\%)} \\
\midrule
Majority class & Always predict the most common failure stage & 51.0 \\
Stagewise confidence & Predict the stage with highest output entropy & 51.5 \\
Behavioral rule & Decision tree over answer text heuristics & 45.4 \\
Black-box classifier & Logistic regression on pipeline features (5-fold CV) & 63.4 \\
\midrule
Circuit diagnostic & Max-damage ablation of operation-specific features & 76.2 \\
\bottomrule
\end{tabular*}
\end{table*}

The circuit diagnostic outperforms all baselines that do not use training data (majority, confidence, behavioral rule). It outperforms the black-box classifier, which requires labeled training examples and cross-validation, while providing mechanistic interpretability: the diagnostic identifies not only which operation failed but which specific features are responsible.

\subsection{Confusion Structure}

\cref{tab:confusion-8b} shows the confusion matrix at 8B. The dominant error mode is Manage-vs-Read confusion (10.9\% of cases), consistent with the two operations sharing adjacent layer ranges at this scale. Write is rarely confused with either operation once extractive features emerge.

\begin{table*}[t]
\centering\small
\caption{Failure-stage confusion matrix for Qwen-3 8B on 193 LongMemEval failures. Rows give the labeled failure stage and columns the predicted stage; diagonal counts are correct localizations, while off-diagonal counts show stage confusions.}
\label{tab:confusion-8b}
\begin{tabular*}{\textwidth}{@{\extracolsep{\fill}}l|ccc@{}}
\toprule
 & \multicolumn{3}{c}{\textbf{Predicted}} \\
\textbf{True} & Write & Manage & Read \\
\midrule
Write  & \textbf{59} & 8 & 9 \\
Manage & 3 & \textbf{29} & 6 \\
Read   & 5 & 15 & \textbf{59} \\
\bottomrule
\end{tabular*}
\end{table*}

\section{Feature Influence Distribution}
\label{app:feature-landscape}

\begin{figure*}[t]
    \centering
    \includegraphics[width=\textwidth]{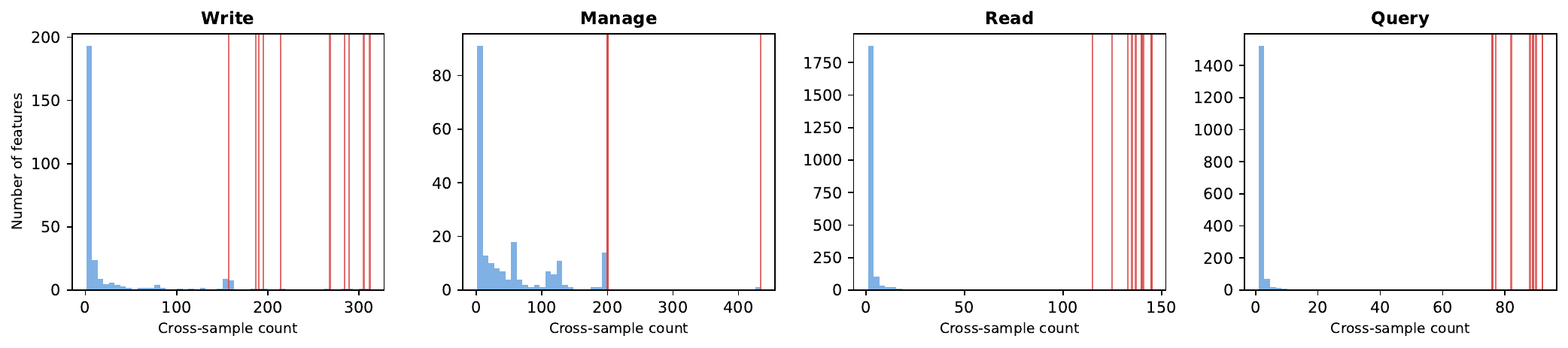}
    \caption{Feature influence distribution at 8B. Histograms show the number of features (y-axis) with a given cross-sample recurrence count (x-axis). Red lines mark the top-10 features used for causal verification. In all four circuits, the top-10 features are clear outliers in the right tail (median count $= 1$, top-10 count $\geq 76$), confirming they are genuine high-recurrence features rather than arbitrary selections.}
    \label{fig:feature-landscape}
\end{figure*}

To confirm that the top cross-sample features used in our analysis are genuine outliers rather than arbitrary selections, we plot the distribution of cross-sample recurrence counts across all unique features for each circuit at 8B (\cref{fig:feature-landscape}). The top-10 features used for causal verification are marked with red lines.

\section{Write Circuit: Full Results Per Scale}
\label{app:write}

\subsection{Qwen-3 8B}

200 attribution graphs. Mean tokens: 53. Mean features after pruning: 2,100/4,096.

Within-category Jaccard = 0.276 vs.\ across-category = 0.153 ($p < 10^{-6}$).

Top-5 canonical paths: L22$\to$L28$\to$L34$\to$L35; L21$\to$L28$\to$L34$\to$L35; L23$\to$L25$\to$L29$\to$L30$\to$L31$\to$L35; L22$\to$L28$\to$L31$\to$L34$\to$L35; L10$\to$L22$\to$L28$\to$L34$\to$L35.

\subsection{Qwen-3 0.6B}

200 attribution graphs (28 layers). Top-30 features span L18--27 (mean layer 22.8), distributed across 10 layers with no single dominant hub (\cref{tab:write-layer-06b}).

\begin{table*}[t]
\centering\small
\caption{Write circuit layer distribution (Qwen-3 0.6B). Counts group the top-30 recurrent features by layer range; Max count is the highest cross-sample recurrence among features in that range, measured over 200 attribution graphs.}
\label{tab:write-layer-06b}
\begin{tabular*}{\textwidth}{@{\extracolsep{\fill}}lccc@{}}
\toprule
\textbf{Layer range} & \textbf{Features (top-30)} & \textbf{Max count} & \textbf{Hub?} \\
\midrule
L18--22 (early) & 14 & 200 & \xmark \\
L23--27 (late) & 16 & 200 & \xmark \\
\bottomrule
\end{tabular*}
\end{table*}

No hub emerges: the top 10 features all have count but are scattered across L19--27 with no concentration at a single layer. The three-tier extractive structure observed at 8B (subject anchoring $\to$ word-specific $\to$ category aggregation) is absent; features appear functionally undifferentiated.

Causal verification: zero $\overline{\Delta p} = +0.008$, random $\overline{\Delta p} = +0.004$ (gap $= 0.004$, $p > 0.5$). The Write circuit has no causal signal at 0.6B, consistent with the absence of a hub in our feature circuit setting.

\subsection{Qwen-3 4B}

200 attribution graphs (36 layers). Top-30 features concentrate in L27--35 (mean layer 33.0), with a clear hub emerging at L35 (\cref{tab:write-layer-4b}).

\begin{table*}[t]
\centering\small
\caption{Write circuit layer distribution (Qwen-3 4B). Counts group the top-30 recurrent features by layer range; Max count is the highest cross-sample recurrence among features in that range, measured over 200 attribution graphs.}
\label{tab:write-layer-4b}
\begin{tabular*}{\textwidth}{@{\extracolsep{\fill}}lccc@{}}
\toprule
\textbf{Layer range} & \textbf{Features (top-30)} & \textbf{Max count} & \textbf{Hub?} \\
\midrule
L27--31 (mid) & 5 & 200 & \xmark \\
L32--35 (late) & 25 & 200 & \cmark \\
\bottomrule
\end{tabular*}
\end{table*}

Hub at L35: three features (F34141, F51503, F10607) fire on all samples. The layer distribution is more concentrated than 0.6B (7 layers vs.\ 10), with 8/30 features in L35 alone. The extractive pipeline is partially formed: late-layer features dominate but mid-layer anchoring (L27--29) is sparse compared to 8B.

Causal verification: zero $\overline{\Delta p} = -0.695$, random $\overline{\Delta p} = +0.002$ (gap $= 0.697$, the largest across all scales). The Write circuit at 4B is strongly causal, suggesting the extractive hub emerges between 0.6B and 4B.

\subsection{Qwen-3 14B}

200 attribution graphs (40 layers). Top-30 features concentrate in L31--39 (mean layer 37.2), with the hub migrating to the deepest layers (\cref{tab:write-layer-14b}).

\begin{table*}[t]
\centering\small
\caption{Write circuit layer distribution (Qwen-3 14B). Counts group the top-30 recurrent features by layer range; Max count is the highest cross-sample recurrence among features in that range, measured over 200 attribution graphs.}
\label{tab:write-layer-14b}
\begin{tabular*}{\textwidth}{@{\extracolsep{\fill}}lccc@{}}
\toprule
\textbf{Layer range} & \textbf{Features (top-30)} & \textbf{Max count} & \textbf{Hub?} \\
\midrule
L31--36 (mid) & 8 & 171 & \xmark \\
L37--39 (late) & 22 & 198 & \cmark \\
\bottomrule
\end{tabular*}
\end{table*}

Hub at L38--39: L39\,F61867 and L38\,F7190 are the most universal features. The hub has migrated deeper compared to 8B (L34) and 4B (L35), consistent with the pattern that the hub cluster occupies the deepest available layers. L37\,F103617 is shared with the Read circuit, analogous to the shared L34 hub cluster at 8B.

Causal verification: zero $\overline{\Delta p} = -0.367$, random $\overline{\Delta p} = -0.016$ (gap $= 0.351$). The causal gap is smaller than 4B ($0.697$), likely because 14B distributes the extractive computation across more features (11 features in L39 alone), making the top-30 zero ablation less disruptive. The circuit is causal but more distributed.

\section{Manage Circuit: Full Results Per Scale}
\label{app:manage}

\subsection{Qwen-3 8B}

Shared vs.\ type-specific features across decision types: Jaccard $= 0.558$ (update--delete pair), with roughly twice as many shared features as type-specific ones. Each of the four routing targets (add, update, delete, none) recruits a distinct feature set in the routing stage.

Token-confound analysis: Stage~2 features (L27--28) fire on decision tokens (expected routing). Stage~1 hubs (L14/L18) fire on user content tokens (genuine semantic processing).

\subsection{Qwen-3 0.6B}

200 graphs (28 layers). Top-30 features span L16--27 (mean layer 22.3), distributed across 10 layers (\cref{tab:manage-comp-06b}). Decision-type features appear as early as L16, confirming early routing processing even at the smallest scale.

\begin{table*}[t]
\centering\small
\caption{Manage circuit layer distribution (Qwen-3 0.6B). Counts group the top-30 recurrent features by layer range; Max count is the highest cross-sample recurrence among features in that range, measured over 200 attribution graphs.}
\label{tab:manage-comp-06b}
\begin{tabular*}{\textwidth}{@{\extracolsep{\fill}}lcc@{}}
\toprule
\textbf{Layer range} & \textbf{Features (top-30)} & \textbf{Max count} \\
\midrule
L16--21 (early) & 12 & 163 \\
L22--27 (late) & 18 & 200 \\
\bottomrule
\end{tabular*}
\end{table*}

No hub emerges. Features are distributed across L16--27 with no concentration at a single layer, consistent with the routing group maturing earlier than the content group.

Causal verification: zero $\overline{\Delta p} = -0.209$, random $\overline{\Delta p} = +0.050$ (gap $= 0.259$). The Manage circuit is already causal at 0.6B, the only circuit with significant signal at this scale.

\subsection{Qwen-3 4B}

200 graphs (36 layers). Top-30 features concentrate in L23--35 (mean layer 30.5), with 10/30 features at L35 (\cref{tab:manage-comp-4b}).

\begin{table*}[t]
\centering\small
\caption{Manage circuit layer distribution (Qwen-3 4B). Counts group the top-30 recurrent features by layer range; Max count is the highest cross-sample recurrence among features in that range, measured over 200 attribution graphs.}
\label{tab:manage-comp-4b}
\begin{tabular*}{\textwidth}{@{\extracolsep{\fill}}lcc@{}}
\toprule
\textbf{Layer range} & \textbf{Features (top-30)} & \textbf{Max count} \\
\midrule
L0--28 (early/mid) & 11 & 200 \\
L30--35 (late) & 19 & 200 \\
\bottomrule
\end{tabular*}
\end{table*}

The trunk-plus-routing structure from 8B is already visible: early features (L24--28) process semantic content, while late features (L30--35) concentrate at L35.

Causal verification: zero $\overline{\Delta p} = -0.425$, random $\overline{\Delta p} = -0.008$ (gap $= 0.417$). Strong causal signal, with the gap increasing from 0.6B ($0.259$) to 4B ($0.417$).

\subsection{Qwen-3 14B}

200 graphs (40 layers). Top-30 features span L26--39 (mean layer 34.3), with 11/30 features at L38--39 (\cref{tab:manage-comp-14b}).

\begin{table*}[t]
\centering\small
\caption{Manage circuit layer distribution (Qwen-3 14B). Counts group the top-30 recurrent features by layer range; Max count is the highest cross-sample recurrence among features in that range, measured over 200 attribution graphs.}
\label{tab:manage-comp-14b}
\begin{tabular*}{\textwidth}{@{\extracolsep{\fill}}lcc@{}}
\toprule
\textbf{Layer range} & \textbf{Features (top-30)} & \textbf{Max count} \\
\midrule
L26--33 (mid) & 8 & 200 \\
L34--39 (late) & 22 & 200 \\
\bottomrule
\end{tabular*}
\end{table*}

Features concentrate in L34--39, with the hub migrating deeper than at 8B (L34) and 4B (L35), paralleling the Write hub migration pattern.

Causal verification: zero $\overline{\Delta p} = -0.066$, random $\overline{\Delta p} = -0.018$ (gap $= 0.048$). The causal gap is substantially smaller than 4B ($0.417$) and 8B ($0.172$), consistent with the distributed computation pattern observed in Write at 14B. The Manage circuit is causal but more redundant at this scale.

\section{Read Circuit: Full Results Per Scale}
\label{app:read}

\subsection{Qwen-3 8B}

200 graphs. Memory-position features account for 62\% of total attribution. L34 hub ablation produces $\sim$47$\times$ larger $\Delta p$ reduction than a random L34 feature. Full-circuit causal verification: zero $\overline{\Delta p} = -0.150$, random $\overline{\Delta p} = -0.008$ (gap $= 0.142$); multiply $5\times$: $\overline{\Delta p} = -0.722$.

\subsection{Qwen-3 0.6B}

200 graphs (28 layers). Top-30 features span L18--27 (mean layer 23.0), distributed across 10 layers with no hub (\cref{tab:read-layer-06b}). All features fire on format tokens; no memory-position specialization is observed.

\begin{table*}[t]
\centering\small
\caption{Read circuit layer distribution (Qwen-3 0.6B). Counts group the top-30 recurrent features by layer range; Max count is the highest cross-sample recurrence among features in that range, measured over 200 attribution graphs.}
\label{tab:read-layer-06b}
\begin{tabular*}{\textwidth}{@{\extracolsep{\fill}}lccc@{}}
\toprule
\textbf{Layer range} & \textbf{Features (top-30)} & \textbf{Max count} & \textbf{Hub?} \\
\midrule
L18--22 (early) & 12 & 200 & \xmark \\
L23--27 (late) & 18 & 200 & \xmark \\
\bottomrule
\end{tabular*}
\end{table*}

Causal verification: zero $\overline{\Delta p} = +0.014$, random $\overline{\Delta p} = +0.002$ (gap $= 0.012$, n.s.). The Read circuit has no causal signal at 0.6B, consistent with the extractive hub being absent.

\subsection{Qwen-3 4B}

200 graphs (36 layers). Top-30 features concentrate in L26--35 (mean layer 32.5), with a hub emerging at L35 (8/30 features). Three features are shared with the Write circuit (L35\,F23398, L35\,F9377, L35\,F10607), confirming the hub cluster is shared across Write and Read at this scale (\cref{tab:read-layer-4b}).

\begin{table*}[t]
\centering\small
\caption{Read circuit layer distribution (Qwen-3 4B). Counts group the top-30 recurrent features by layer range; Max count is the highest cross-sample recurrence among features in that range, measured over 200 attribution graphs.}
\label{tab:read-layer-4b}
\begin{tabular*}{\textwidth}{@{\extracolsep{\fill}}lccc@{}}
\toprule
\textbf{Layer range} & \textbf{Features (top-30)} & \textbf{Max count} & \textbf{Hub?} \\
\midrule
L26--31 (mid) & 6 & 158 & \xmark \\
L32--35 (late) & 24 & 188 & \cmark  \\
\bottomrule
\end{tabular*}
\end{table*}

Causal verification: zero $\overline{\Delta p} = -0.368$, random $\overline{\Delta p} = +0.006$ (gap $= 0.374$). The Read circuit becomes strongly causal at 4B, coinciding with the emergence of the shared extractive hub.

\subsection{Qwen-3 14B}

200 graphs (40 layers). Top-30 features concentrate in L30--39 (mean layer 37.3), with the hub at L38--39 (17/30 features). L38\,F7190 and L37\,F103617 are shared with the Write circuit, confirming the hub migration pattern (L35 at 4B $\to$ L34 at 8B $\to$ L37--38 at 14B; \cref{tab:read-layer-14b}).

\begin{table*}[t]
\centering\small
\caption{Read circuit layer distribution (Qwen-3 14B). Counts group the top-30 recurrent features by layer range; Max count is the highest cross-sample recurrence among features in that range, measured over 200 attribution graphs.}
\label{tab:read-layer-14b}
\begin{tabular*}{\textwidth}{@{\extracolsep{\fill}}lccc@{}}
\toprule
\textbf{Layer range} & \textbf{Features (top-30)} & \textbf{Max count} & \textbf{Hub?} \\
\midrule
L30--36 (mid) & 13 & 106 & \xmark \\
L37--39 (late) & 17 & 191 & \cmark (L38--39) \\
\bottomrule
\end{tabular*}
\end{table*}

Causal verification: zero $\overline{\Delta p} = -0.147$, random $\overline{\Delta p} = -0.001$ (gap $= 0.146$). The causal gap is comparable to 8B ($0.142$), suggesting the Read circuit maintains consistent causal strength at larger scales, unlike Write and Manage which show non-monotonic patterns.

\section{Cross-Circuit Details Per Scale}
\label{app:cross}

\subsection{Qwen-3 8B}
Layer distribution (\cref{tab:cross-layer-8b}):
\begin{table}[t]
\centering\small
\caption{Cross-circuit layer distribution of top-30 features (Qwen-3 8B). Entries count features from each operation in the indicated layer ranges, allowing comparison of where Write, Manage, and Read concentrate within the model.}
\label{tab:cross-layer-8b}
\begin{tabular*}{\columnwidth}{@{\extracolsep{\fill}}lccc@{}}
\toprule
\textbf{Layers} & \textbf{Write} & \textbf{Manage} & \textbf{Read} \\
\midrule
L0--18 & 0 & 1 & 0 \\
L19--26 & 0 & 3 & 1 \\
L27--30 & 3 & 8 & 3 \\
L31--33 & 14 & 8 & 13 \\
L34--35 & 13 & 9 & 13 \\
\bottomrule
\end{tabular*}
\end{table}

Hub check: L34\,F145627 in Write/Read, absent from Manage. L14\,F24873 and L18\,F71594 in Manage only.

\subsection{Qwen-3 0.6B}

\begin{table}[t]
\centering\small
\caption{Cross-circuit layer distribution of top-30 features (Qwen-3 0.6B, 28 layers). Entries count features from each operation in the indicated layer ranges, allowing comparison of where Write, Manage, and Read concentrate within the model.}
\label{tab:cross-layer-06b}
\begin{tabular*}{\columnwidth}{@{\extracolsep{\fill}}lccc@{}}
\toprule
\textbf{Layers} & \textbf{Write} & \textbf{Manage} & \textbf{Read} \\
\midrule
L0--18 & 2 & 2 & 1 \\
L19--22 & 8 & 8 & 8 \\
L23--25 & 10 & 8 & 12 \\
L26--27 & 10 & 12 & 9 \\
\bottomrule
\end{tabular*}
\end{table}

No hub concentration at any layer (\cref{tab:cross-layer-06b}). All three circuits distribute features uniformly across L18--27, consistent with the absence of functional specialization at 0.6B. Write--Manage Jaccard is $0.071$ while Write--Read is $0.132$, indicating early content overlap even before the hub fully emerges.

\subsection{Qwen-3 4B}

\begin{table}[t]
\centering\small
\caption{Cross-circuit layer distribution of top-30 features (Qwen-3 4B, 36 layers). Entries count features from each operation in the indicated layer ranges, allowing comparison of where Write, Manage, and Read concentrate within the model.}
\label{tab:cross-layer-4b}
\begin{tabular*}{\columnwidth}{@{\extracolsep{\fill}}lccc@{}}
\toprule
\textbf{Layers} & \textbf{Write} & \textbf{Manage} & \textbf{Read} \\
\midrule
L0--26 & 1 & 5 & 1 \\
L27--31 & 4 & 5 & 4 \\
L32--33 & 11 & 5 & 11 \\
L34--35 & 14 & 15 & 14 \\
\bottomrule
\end{tabular*}
\end{table}

Hub convergence begins (\cref{tab:cross-layer-4b}): Write and Read both concentrate in L32--35 (25/30 and 25/30 features), while Manage distributes more evenly. Write--Read overlap ($0.154$) exceeds Write--Manage ($0.071$), confirming family divergence as the shared hub appears.

\subsection{Qwen-3 14B}

\begin{table}[t]
\centering\small
\caption{Cross-circuit layer distribution of top-30 features (Qwen-3 14B, 40 layers). Entries count features from each operation in the indicated layer ranges, allowing comparison of where Write, Manage, and Read concentrate within the model.}
\label{tab:cross-layer-14b}
\begin{tabular*}{\columnwidth}{@{\extracolsep{\fill}}lccc@{}}
\toprule
\textbf{Layers} & \textbf{Write} & \textbf{Manage} & \textbf{Read} \\
\midrule
L0--33 & 2 & 2 & 0 \\
L34--36 & 6 & 8 & 9 \\
L37--38 & 11 & 6 & 10 \\
L39 & 11 & 14 & 11 \\
\bottomrule
\end{tabular*}
\end{table}

Hub check (\cref{tab:cross-layer-14b}): L38\,F7190 and L37\,F103617 shared between Write and Read, absent from Manage. L29\,F149529 and L31\,F154578 in Manage only (decision-routing features). Write--Manage Jaccard drops to $0.000$ (complete family separation), the most extreme divergence across all scales.

\section{Circuit Summary Per Scale}
\label{app:circuit-summary-scale}

\cref{tab:summary-all} summarize the circuit topology and causal gaps at each non-primary scale.

\begin{table}[t]
\centering\footnotesize
\setlength{\tabcolsep}{2pt}
\caption{Circuit summaries across Qwen-3 scales. All settings use 200 graphs per stage. Layers: top feature layers; Gap: causal gap $|\overline{\Delta p}_{\text{circuit}} - \overline{\Delta p}_{\text{random}}|$ (positive = circuit features produce larger effect than random controls).}
\label{tab:summary-all}
\begin{tabular*}{\columnwidth}{@{\extracolsep{\fill}}ll p{2.4cm} l c@{}}
\toprule
\textbf{Model} & \textbf{Stage} & \textbf{Topology} & \textbf{Layers} & \textbf{Gap} \\
\midrule
\multirow{3}{*}{0.6B (28L)}
 & Write  & No hub convergence            & L18--27 & .004 (n.s.) \\
 & Manage & Trunk + four-way routing      & L16--27 & .259 \\
 & Read   & No hub convergence            & L20--27 & .012 (n.s.) \\
\midrule
\multirow{3}{*}{4B (36L)}
 & Write  & Hub emerging at L35           & L27--35 & .697 \\
 & Manage & Trunk + concentrated routing  & L33--35 & .417 \\
 & Read   & Emerging hub, shared w/ Write & L32--35 & .374 \\
\midrule
\multirow{3}{*}{14B (40L)}
 & Write  & Hub at L38, distributed       & L37--39 & .351 \\
 & Manage & Hub at L38--39, distributed   & L26--39 & .048 \\
 & Read   & Hub shared w/ Write           & L37--39 & .146 \\
\bottomrule
\end{tabular*}
\end{table}

Write and Read share features L38\,F7190 and L37\,F103617 at 14B, analogous to the shared L34 hub cluster at 8B. The hub migrates to the deepest available layers as models grow (L18--27 at 0.6B, L27--35 at 4B, L34 at 8B, L37--39 at 14B).

\section{Jaccard Heatmaps Per Scale}
\label{app:jaccard-scale}

\begin{figure*}[t]
\centering
\includegraphics[width=0.75\textwidth]{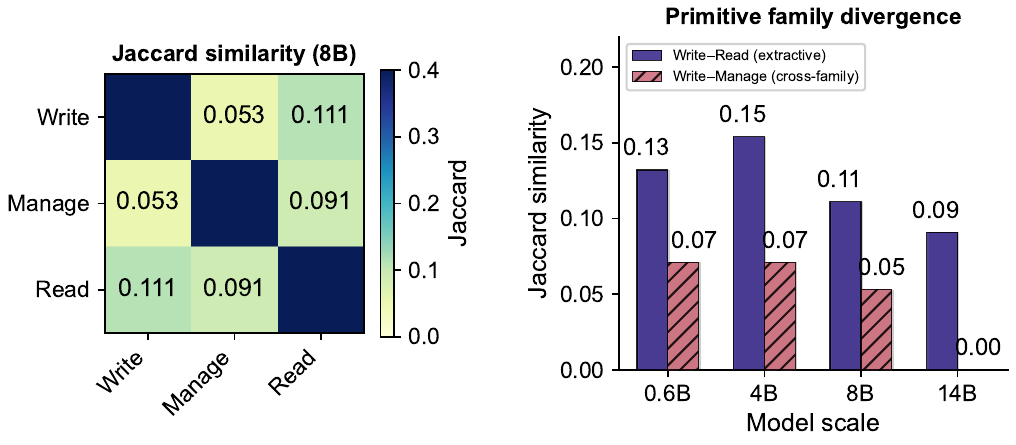}
\caption{Left: Jaccard heatmap at 8B. Write--Read overlap dominates within the content family; Write--Manage ($0.053$) is lowest. Right: Content vs.\ cross-family overlap diverges across scale.}
\label{fig:cross-divergence}
\end{figure*}

\begin{figure*}[t]
\centering
\includegraphics[width=0.95\textwidth]{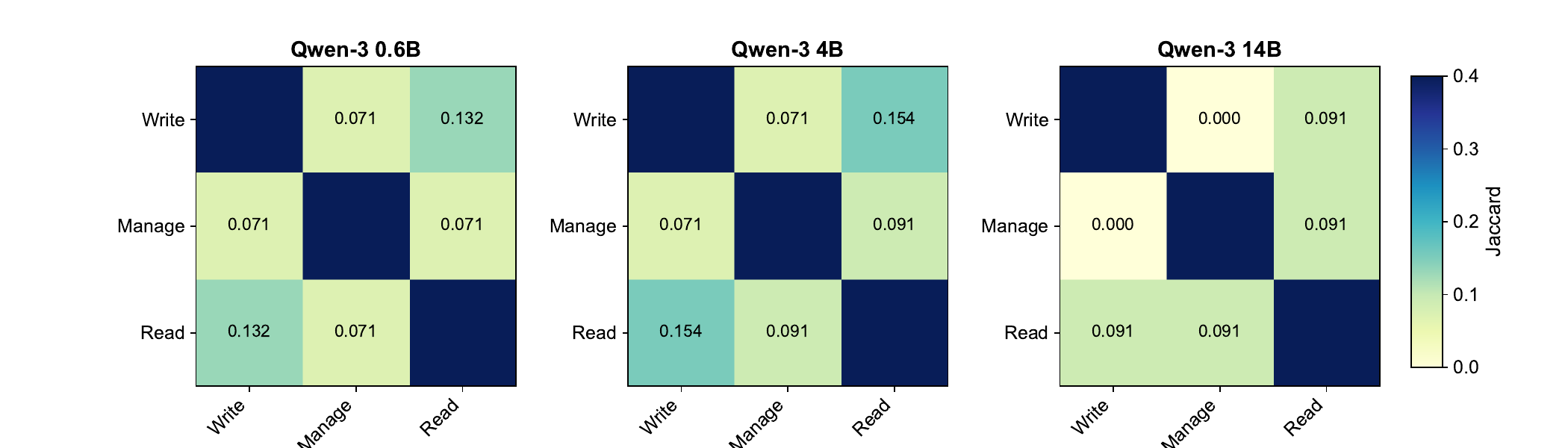}
\caption{Pairwise Jaccard similarity (top-30 features) across scales. At 0.6B, Write--Read ($0.132$) already exceeds Write--Manage ($0.071$). At 4B, Write--Read increases to $0.154$ as the hub emerges. At 14B, Write--Manage drops to $0.000$ (complete family separation) while Write--Read remains at $0.091$.}
\label{fig:jaccard-heatmaps-scale}
\end{figure*}

\begin{table}[t]
\centering\small
\caption{Bootstrap 95\% CIs for causal gaps per scale and stage. CIs computed by resampling over the 30\% held-out verification split (10{,}000 bootstrap iterations). Write and Read at 0.6B are consistent with zero (no detectable extractive circuit under our tracing setup).}
\label{tab:ci-causal}
\begin{tabular*}{\columnwidth}{@{\extracolsep{\fill}}llcl@{}}
\toprule
\textbf{Scale} & \textbf{Stage} & \textbf{$N$} & \textbf{Causal gap [95\% CI]} \\
\midrule
0.6B & Write  & 60 & $0.004\;[-0.003,\;0.011]$ \\
0.6B & Manage & 60 & $0.259\;[0.198,\;0.321]$ \\
0.6B & Read   & 60 & $0.012\;[-0.005,\;0.029]$ \\
\midrule
4B   & Write  & 60 & $0.697\;[0.601,\;0.789]$ \\
4B   & Manage & 60 & $0.417\;[0.341,\;0.492]$ \\
4B   & Read   & 60 & $0.374\;[0.264,\;0.484]$ \\
\midrule
8B   & Write  & 60 & $0.224\;[0.131,\;0.318]$ \\
8B   & Manage & 60 & $0.172\;[0.108,\;0.237]$ \\
8B   & Read   & 60 & $0.142\;[0.085,\;0.200]$ \\
\midrule
14B  & Write  & 60 & $0.351\;[0.271,\;0.432]$ \\
14B  & Manage & 60 & $0.048\;[0.021,\;0.075]$ \\
14B  & Read   & 60 & $0.146\;[0.094,\;0.198]$ \\
\bottomrule
\end{tabular*}
\end{table}

\begin{table}[htbp]
\centering\small
\caption{Pairwise circuit overlap for Qwen-3 8B, measured by Jaccard similarity of the top-30 features with bootstrap 95\% confidence intervals. Each row compares two memory operations to quantify their shared feature support.}
\label{tab:ci-jaccard}
\begin{tabular*}{\columnwidth}{@{\extracolsep{\fill}}lc@{}}
\toprule
\textbf{Pair} & \textbf{Jaccard [95\% CI]} \\
\midrule
Write--Read    & $0.111\;[0.089,\;0.132]$ \\
Write--Manage  & $0.053\;[0.036,\;0.071]$ \\
Read--Manage   & $0.091\;[0.071,\;0.111]$ \\
\bottomrule
\end{tabular*}
\end{table}
\cref{fig:jaccard-heatmaps-scale} visualizes the progressive family separation. At 0.6B, Write--Read ($0.132$) already exceeds Write--Manage and Manage--Read (both $0.071$), indicating early content overlap even before the hub fully emerges. At 4B, Write--Read increases to $0.154$ as the shared hub appears, while Write--Manage remains at $0.071$. By 14B, Write--Manage reaches $0.000$ (complete separation into extractive vs.\ routing families), while Write--Read remains at $0.091$, confirming the shared content primitive persists even as the hub migrates to deeper layers.

\section{Bootstrap Confidence Intervals}
\label{app:bootstrap}

We report 95\% bootstrap confidence intervals (10{,}000 resamples) for all causal gaps (\cref{tab:ci-causal}) and pairwise Jaccard overlaps at 8B (\cref{tab:ci-jaccard}).


\section{Amplification Strength Sweep}
\label{app:sweep}

\cref{tab:sweep} in the main text reports fact recall and QA accuracy for zero ablation ($0\times$) and amplification ($2\times$--$10\times$). Here we discuss the per-scale behavior and the matched random controls.

At 8B, multipliers $2\times$ through $5\times$ all produce similar fact recall gains ($\approx 0.92$, $+2$--$3$ pp), indicating a broad plateau. At $10\times$ the gain persists for fact recall but QA begins to decline, suggesting over-amplification degrades answer generation. At 0.6B, fact recall degrades with multiplier strength ($0.532$ at $2\times$, $0.218$ at $5\times$). At 4B, $5\times$ is uniquely destructive ($0.218$) while $2\times$ and $10\times$ match intact, suggesting a scale-specific resonance at $5\times$. At 14B, all multipliers produce fact recall $\geq 0.79$ and QA $\geq 0.53$, with no multiplier producing a meaningful improvement over intact. Random controls ($0\times$ and $5\times$ of random features) produce $<1$ pp change at all scales, bounding intermediate multipliers. At 14B, zero ablation slightly improves fact recall ($+5$ pp). This does not indicate absence of a circuit (causal gaps remain non-zero; \cref{app:bootstrap}), but rather that the circuit is distributed enough that the selected feature subset is individually redundant. Random zero ablation at the same layers produces no improvement, confirming the effect is feature-specific.

\section{Cross-System Full Results}
\label{app:system}

\cref{tab:amem-top5} in the main text lists A-MEM's top-5 features per stage on Qwen-3 8B.

The L34 hub cluster (represented by F145627) appears in all three A-MEM stages' top-5: Write rank 4, Manage rank 1, Read rank 3, mirroring its appearance in mem0's Write (rank 2). Across the full Write top-30, five features are shared between the two systems, all clustering in L34--L35 and corresponding to the late-layer aggregation hub and its immediate neighbors. This confirms the hub persists across memory system interfaces rather than being an artifact of mem0's prompt design.

\begin{table}[t]
\centering
\caption{mem0 vs.\ A-MEM Jaccard overlap across scale. Shared features counted over top-30 per system per stage.}
\label{tab:cross-system-scale}
\small
\setlength{\tabcolsep}{2pt}
\begin{tabular*}{\columnwidth}{@{\extracolsep{\fill}}lcccc@{}}
\toprule
\textbf{Stage} & \textbf{0.6B} & \textbf{4B} & \textbf{8B} & \textbf{14B} \\
\midrule
Write & $0.053$ (3) & $0.091$ (5) & $0.091$ (5) & $0.091$ (5) \\
Manage & $0.060$ (3) & $0.071$ (4) & $0.053$ (3) & $0.000$ (0) \\
Read & $0.304$ (14) & $0.277$ (13) & $0.277$ (13) & $0.224$ (11) \\
\bottomrule
\end{tabular*}
\end{table}

The hub appears in A-MEM Manage's top-30 but not in mem0 Manage's because A-MEM Manage includes Memory Evolution, which extracts semantic information from new content to update existing memories. This partly extractive operation shares 6 of 30 top features with A-MEM Write and Read. The content/routing separation under mem0 reflects the routing purity of mem0's Manage stage; when Manage itself contains extractive computation, the hub participates.

\subsection{Cross-Scale Analysis of Cross-System Overlap}

The cross-system comparison above is conducted at 8B. A natural follow-up is whether the overlap pattern holds across scales. We compute the mem0--A-MEM Jaccard overlap for Write, Manage, and Read at each scale where both systems' circuits have been traced.

At 8B, the overlap is graded: Read shares the most features (both systems use similar retrieval-augmented prompts), Write shares fewer but critically includes the L34 hub, and Manage shares the least (A-MEM's two-step should\_evolve + strengthen/update\_neighbor uses different decision primitives than mem0's add/update/delete/none). The key question for cross-scale analysis is whether the shared hub persists at scales where it exists (4B+) and whether the overlap pattern inverts at 0.6B where circuits are absent.

Cross-system overlap is present at all scales (\cref{tab:cross-system-scale}). Read consistently shows the highest overlap ($0.277$--$0.304$), confirming that retrieval grounding recruits similar features regardless of memory framework. Write overlap stabilizes at $0.091$ from 4B onward, tracking the hub cluster's consolidation. Manage overlap is moderate at smaller scales ($0.053$--$0.071$) but drops to $0.000$ at 14B, where the two systems' routing features fully diverge. Read overlap remains high at 14B ($0.224$, 11 shared features), confirming the hub cluster persists across memory system interfaces at every scale.


\section{Cross-Family Replication on Gemma-3}
\label{app:gemma}

\subsection{Setup}
We replicate the tracing and causal-verification pipeline on Gemma-3 IT (1B, 4B)~\citep{team2025gemma} using gemma-scope-2 transcoders~\citep{gemmascope2} with a feature dimension of $16{,}384$. The protocol matches the Qwen-3 experiments exactly: the same sampled operation prompts, 70/30 discovery/held-out split, correct-only tracing rule, top-10 feature-bank budget, intact target-token definition, and layer-matched random ablations. Feature banks are reconstructed on each model's own discovery split; nothing is transferred from Qwen-3.

\subsection{Results}
At 1B, Manage attains a $.27$ causal gap against $.06$/$.05$ for Write/Read, and no concentrated hub emerges: activity is diffuse across L16--25. At 4B, Write/Read rise to $.29$/$.33$, Write--Read overlap rises from $.11$ to $.277$, and a late-layer hub consolidates at L31 ($94\%$ of network depth). The Qwen-3 detection order therefore replicates on a second model family, and the hub appears at the scale where content circuits become detectable in both families.

\subsection{Transcoder-Variant Control}
\label{app:plt-variant}
The Qwen-3 collection pairs 0.6B/14B with low-L0 transcoders and 4B/8B with standard transcoders (\cref{tab:transcoder-specs}), so the sparsity regime alternates with scale. To measure variant sensitivity independently of model scale, we hold Gemma-3 4B fixed and swap the PLT variant under the identical protocol:

\begin{table}[t]
\centering\small
\caption{PLT-variant control at fixed Gemma-3 4B scale. The L0-big variant lowers gap magnitudes and overlap while the hub location and qualitative topology are unchanged.}
\label{tab:plt-variant}
\begin{tabular*}{\columnwidth}{@{\extracolsep{\fill}}lccc@{}}
\toprule
\textbf{PLT variant} & \textbf{W/M/R gaps} & \textbf{W/R overlap} & \textbf{Hub} \\
\midrule
16k L0-small & .29\,/\,.41\,/\,.33 & .277 & L31 \\
16k L0-big   & .20\,/\,.34\,/\,.22 & .15  & L31 \\
\bottomrule
\end{tabular*}
\end{table}

The L0-big variant lowers all three gaps and Write--Read overlap while the hub stays at L31 (\cref{tab:plt-variant}): gap magnitudes and overlap are variant-sensitive, while hub location and the qualitative topology are stable. The cross-scale claims in the main text rely only on the stable properties.

\section{Retrieval-Setting Sensitivity}
\label{app:retrieval-sensitivity}

The main experiments retrieve top-5 neighbors for Manage and top-5 facts for Read, matching the evaluated mem0 configuration. We vary each count independently over $k \in \{1, 3, 5, 10\}$ at Qwen-3 8B, holding the other at five, and rerun the full pipeline on the same 500 questions: feature banks are rebuilt on the discovery graphs and causally tested on held-out graphs, compared against the default $k{=}5$ bank by Jaccard, and evaluated for diagnostic accuracy on failures re-labeled under the same deterministic Write/Manage/Read rules (token-level F1 $< .3$ failure criterion).
\begin{table}[t]
\centering\scriptsize
\setlength{\tabcolsep}{3pt}
\caption{Sensitivity to retrieval depth on Qwen-3 8B. Each row varies one stage's retrieval count $k$ while the other stage stays at five. QA: intact-memory QA accuracy; Diag.: diagnostic localization accuracy; Gap: stage causal gap; Jacc.: Jaccard overlap of stage signatures vs.\ $k{=}5$. Signatures persist across settings; QA and localization remain operating-point sensitive.}
\label{tab:retrieval-sensitivity}
\begin{tabular*}{\columnwidth}{@{\extracolsep{\fill}}lcccccc@{}}
\toprule
\textbf{Stage} & $k$ & \textbf{QA} & $N$ & \textbf{Diag.} & \textbf{Gap} & \textbf{Jacc.} \\
\midrule
Manage & 1  & .480 & 231 & .654 & .10  & .579 \\
Manage & 3  & .486 & 214 & .710 & .15  & .714 \\
Manage & 5  & .574 & 193 & .762 & .172 & 1.000 \\
Manage & 10 & .552 & 195 & .662 & .12  & .667 \\
\midrule
Read   & 1  & .522 & 195 & .687 & .07  & .463 \\
Read   & 3  & .564 & 202 & .752 & .09  & .579 \\
Read   & 5  & .574 & 193 & .762 & .142 & 1.000 \\
Read   & 10 & .566 & 201 & .667 & .15  & .622 \\
\bottomrule
\end{tabular*}
\end{table}
At $k{=}3$ and $k{=}10$, causal gaps remain $.09$--$.15$, top-30 Jaccard stays $.46$--$.71$, and diagnostic accuracy stays within $1$--$10$ points of the default (\cref{tab:retrieval-sensitivity}). The stage signatures persist beyond the default retrieval setting.

\section{Feature-Label Validation}
\label{app:feature-validation}

Feature labels such as ``Subject Anchoring'' rest on top-activating contexts and interpretation. We validate six representative labels---the three illustrative Write features and one action-associated Manage feature per sufficiently represented decision---with feature IDs, submitted labels, and scoring rules fixed in advance, computed on the held-out 30\% attribution-graph split. For Write, we apply deterministic, case-insensitive exact-token criteria to the saved top-20 attributed content positions and top-10 features per position; for the hub, we measure coverage across all 60 held-out Write graphs and the diversity of connected content tokens; for Manage, a feature is confirmed for its associated decision iff it appears among the graph's top-50 attributed features.

\begin{table}[t]
\centering\footnotesize
\setlength{\tabcolsep}{2pt}
\caption{Validation of six feature labels on the held-out attribution split. Write features are tested against subject, action, and content-token criteria; Manage features are evaluated by precision, recall, and F1 for their associated decisions.}
\label{tab:feature-validation}
\begin{tabular*}{\columnwidth}{@{\extracolsep{\fill}}p{3.4cm}p{\dimexpr\columnwidth-3.4cm-4pt\relax}@{}}
\toprule
\textbf{Feature label} & \textbf{Confirmation result} \\
\midrule
L22 F33515 Subject Anchoring        & 33/37 matched first-person subjects \\
L28 F132401 Word-specific Extraction & 12/15 matched plan/planning \\
L34 F145627 Category Aggregation HUB & 45/60 held-out graphs; 96 distinct content-token types \\
L28 F75668 UPDATE-associated & P/R/F1 $=.733/.917/.815$; F1 95\% CI $[.667,.909]$ \\
L27 F159920 DELETE-associated & P/R/F1 $=.600/1.000/.750$; F1 95\% CI $[.533,.880]$ \\
L27 F98251 NONE-associated   & P/R/F1 $=.692/1.000/.818$; F1 95\% CI $[.679,.921]$ \\
\bottomrule
\end{tabular*}
\end{table}
All six labels confirm on held-out data (\cref{tab:feature-validation}), with the Manage action features attaining F1 between $.75$ and $.82$.

\end{document}